\documentclass[10pt,twocolumn,letterpaper]{article}

\usepackage{cvpr}

 \usepackage[sectionbib]{bibunits}

\definecolor{cvprblue}{rgb}{0.21,0.49,0.74}
\usepackage[pagebackref,breaklinks,colorlinks,allcolors=cvprblue]{hyperref}
\usepackage{amssymb,amsmath,amsthm,graphicx,color,hyperref,enumerate}
\usepackage[ruled, linesnumbered, noend]{algorithm2e}
\SetKwComment{Comment}{\% }{}
\usepackage{bibunits}
\usepackage{array}
\newcolumntype{H}{>{\setbox0=\hbox\bgroup}c<{\egroup}@{}}

\def\paperID{***}
\def\confName{CVPR}
\def\confYear{2026}
\def\MethodName{GeodesicNVS}
\newtheorem{theorem}{Theorem}
\newtheorem{corollary}[theorem]{Corollary}
\newtheorem{lemma}[theorem]{Lemma}
\newtheorem{proposition}[theorem]{Proposition}
\newtheorem{axiom}[theorem]{Axiom}
\newtheorem{definition}[theorem]{Definition}
\newtheorem{remark}[theorem]{Remark}
\newtheorem{notation}[theorem]{Notation}
\newtheorem{example}[theorem]{Example}

\newcommand{\be}{\begin{enumerate}}
\newcommand{\ee}{\end{enumerate}}
\newcommand{\bi}{\begin{itemize}}
\newcommand{\ei}{\end{itemize}}
\newcommand{\ba}{\begin{array}}
\newcommand{\ea}{\end{array}}
\newcommand{\bt}{\begin{tabular}}
\newcommand{\et}{\end{tabular}}
\newcommand{\btb}{\begin{tabbing}}
\newcommand{\etb}{\end{tabbing}}
\newcommand{\bfg}{\begin{figure}}
\newcommand{\efg}{\end{figure}}
\newcommand{\bsl}{\begin{slide}}
\newcommand{\esl}{\end{slide}}
\newcommand{\bthm}{\begin{theorem}}
\newcommand{\ethm}{\end{theorem}}
\newcommand{\bcor}{\begin{corollary}}
\newcommand{\ecor}{\end{corollary}}
\newcommand{\blem}{\begin{lemma}}
\newcommand{\elem}{\end{lemma}}
\newcommand{\bprop}{\begin{proposition}}
\newcommand{\eprop}{\end{proposition}}
\newcommand{\basm}{\begin{assumption}}
\newcommand{\easm}{\end{assumption}}
\newcommand{\baxm}{\begin{axiom}}
\newcommand{\eaxm}{\end{axiom}}
\newcommand{\bdfn}{\begin{definition}}
\newcommand{\edfn}{\end{definition}}
\newcommand{\brmk}{\begin{remark}}
\newcommand{\ermk}{\end{remark}}
\newcommand{\balg}{\begin{algorithm}}
\newcommand{\ealg}{\end{algorithm}}
\newcommand{\bntn}{\begin{notation}}
\newcommand{\entn}{\end{notation}}
\newcommand{\bexm}{\begin{example}}
\newcommand{\eexm}{\end{example}}
\newcommand{\bpf}{\begin{proof}}
\newcommand{\epf}{\end{proof}}

\newcommand{\cK}{\mathcal{K}}

\renewcommand{\hat}{\widehat}
\renewcommand{\tilde}{\widetilde}

\newcommand{\transpose}{^\mathsf{T}}
\newcommand{\gd}{\dot{\gamma}}
\newcommand{\gdd}{\ddot{\gamma}}

\newcommand{\gdh}{\hat{\dot{\gamma}}}
\newcommand{\dd}[2]{\frac{\text{d} #1}{\text{d} #2}}

\newcommand{\dt}{\text{ d}t}

\newcommand{\ddelta}[2]{\frac{\delta #1}{\delta #2}}

\usepackage{xspace}
\makeatletter
\DeclareRobustCommand\onedot{\futurelet\@let@token\@onedot}
\def\@onedot{\ifx\@let@token.\else.\null\fi\xspace}

\title{\MethodName: Probability Density Geodesic Flow Matching

for Novel View Synthesis}

\author{
Xuqin Wang\textsuperscript{1,2} \quad
Tao Wu\textsuperscript{1} \quad
Yanfeng Zhang\textsuperscript{1} \quad
Lu Liu\textsuperscript{1} \quad
Mingwei Sun\textsuperscript{1} \quad
Yongliang Wang\textsuperscript{1} \\
Niclas Zeller\textsuperscript{3} \quad
Daniel Cremers\textsuperscript{2} \\
\textsuperscript{1}Huawei Hilbert Research Center (Dresden) \\
\textsuperscript{2}Technical University of Munich \\
\textsuperscript{3}Karlsruhe University of Applied Sciences\\
}

\begin{document}
\maketitle
\begin{abstract}

Recent advances in generative modeling have substantially enhanced novel view synthesis, yet maintaining consistency across viewpoints remains challenging. Diffusion-based models rely on stochastic noise-to-data transitions, which obscure deterministic structures and yield inconsistent view predictions.
We advocate a Data-to-Data Flow Matching framework that learns deterministic transformations between paired views, enhancing view-consistent synthesis through explicit data coupling.
Building on this, we propose Probability Density Geodesic Flow Matching (PDG-FM), which aligns interpolation trajectories with density-based geodesics of a data manifold. To enable tractable geodesic estimation, we employ a teacher-student framework that distills density-based geodesic interpolants into an efficient ambient-space predictor.
Empirically, our method surpasses diffusion-based baselines on Objaverse and GSO30 datasets, demonstrating improved structural coherence and smoother transitions across views. These results highlight the advantages of incorporating data-dependent geometric regularization into deterministic flow matching for consistent novel view generation.

\end{abstract}    
\section{Introduction}
\label{sec:intro}

Novel View Synthesis (NVS) aims to generate unseen views of a scene from a limited set of observations. Despite the rapid progress of generative models~\cite{rombach2022high, podell2023sdxlimprovinglatentdiffusion}, achieving consistent and geometry-aware view generation remains challenging.
Recent research has significantly enhanced the controllability and structural reasoning of diffusion models through latent alignment and verifiable consistency mechanisms~\cite{wang2025ladb, zhou2026unified, long2026spatialreward, chen2026driving, song2025hume, dai2025latent, dai2026omni2sound}.
However, achieving consistent viewpoint synthesis remains challenging, as most generative models lack explicit modeling of the underlying geometric relationships between views~\cite{poole2022dreamfusiontextto3dusing2d,tang2023makeit3dhighfidelity3dcreation}.

Flow Matching (FM) \cite{lipman2023flow,liu2023flow,albergo2023building} provides a deterministic alternative, learning continuous-time dynamics between data distributions. Conditional Flow Matching (CFM)~\cite{lipman2023flow,liu2023flow,albergo2023building} extends this concept to conditional distributions, enabling mappings between structured data pairs such as different views of an object. 
However, most current formulations of CFM rely on simple linear interpolants between source and target data~\cite{Schusterbauer2024,chadebec2025lbmlatentbridgematching,fischer2025flowr}, which may not faithfully capture the nonlinear geometry of the data manifold in latent space, potentially leading to suboptimal transitions between viewpoints.

\begin{figure*}[t!]\centering
\begin{minipage}{0.23\linewidth}
    \includegraphics[width=\linewidth]{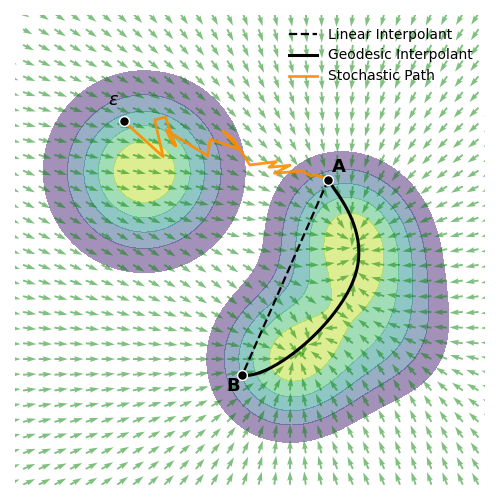}
\end{minipage}
\hfill
\begin{minipage}{0.72\linewidth}
\raisebox{0.5em}{\rotatebox{90}{\scriptsize Diffusion}}\hfill
\includegraphics[width=0.95\linewidth]{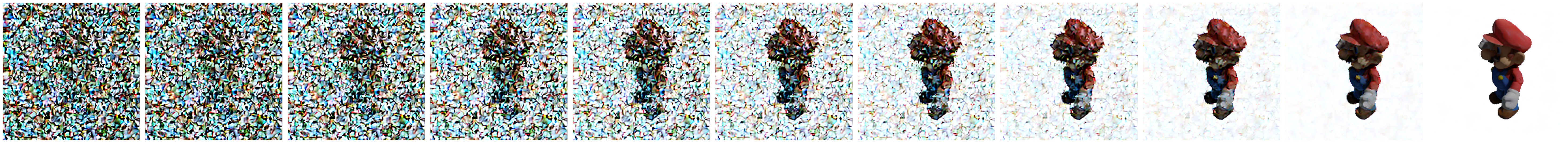}\vfill
\raisebox{0.5em}{\rotatebox{90}{\scriptsize Linear}}\hfill
\includegraphics[width=0.95\linewidth]{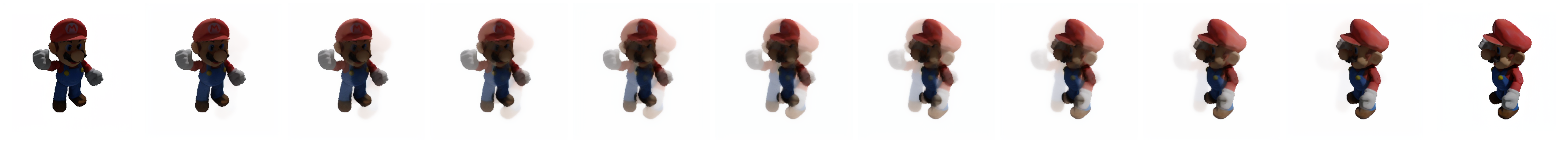}\vfill
\raisebox{0.5em}{\rotatebox{90}{\scriptsize Geodesic}}\hfill
\includegraphics[width=0.95\linewidth]{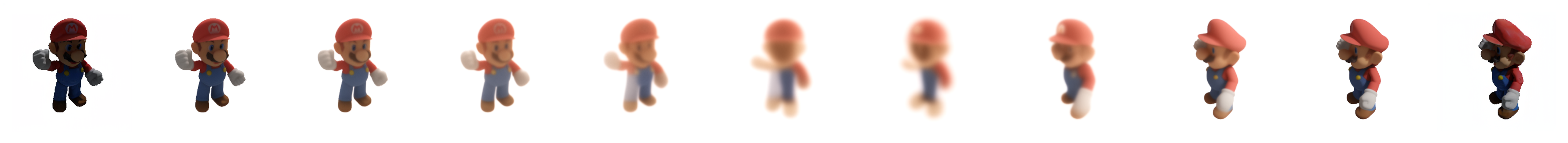}
\end{minipage}
\caption{
\textbf{From Conditional Diffusion Model to Probability Density Geodesic Flow Matching.}
Conventional diffusion models learn stochastic noise-to-data transitions, often losing deterministic structure.
We instead train a Data-to-Data Flow Matching network to learn continuous deterministic transformations between paired data samples $(x_0, x_1)$.
To ensure geometric consistency, we propose to deploy a probability-density-based geodesic to align flow matching interpolants with high-density regions of the data manifold.
This unified design couples accurate data coupling with manifold-aware regularization, 
yielding realistic, view-consistent transformations.
}
\vspace{-.3cm}
\label{fig:teaser}
\end{figure*}

To address this, we propose Probability Density Geodesic Flow Matching (PDG-FM), a framework that integrates data-dependent geometric regularization into conditional flow matching.
Built upon the Data-to-Data Flow Matching framework, our method learns deterministic flows between paired samples $(x_0, x_1)$, preserving structural correspondence, rather than relying on noise-to-data transitions.
We further refine these flows using a probability-density-based geodesic optimization, which aligns flow matching interpolants $x_t$ with the data manifold.
The local metric is defined inversely proportional to the learned data density, encouraging paths that traverse high-probability regions while penalizing off-manifold deviations.
As shown in Fig.~\ref{fig:teaser}, GeodesicNet learns smooth, energy-stable interpolants that respect the underlying geometry.

Empirically, PDG-FM improves both perceptual quality and viewpoint consistency in novel view synthesis.
Compared to diffusion and standard flow matching baselines, it produces view-consistent transformations that remain faithful to the input geometry while retaining visual realism.
Our analysis further shows that the resulting geodesic interpolants exhibit higher optical flow magnitude and lower Euler--Lagrange residuals, confirming their adherence to meaningful manifold structure.

Overall, our contributions are threefold:
\begin{itemize}
  \item We formulate a Data-to-Data Flow Matching framework as an alternative to diffusion-based conditional modeling for paired-view NVS, and instantiate its linear-interpolant baseline as Linear-D2D-FM.
  \item We introduce Probability Density Geodesic Flow Matching (PDG-FM) with a teacher-student distillation framework, where a teacher estimates density-based geodesic interpolants and distills them into GeodesicNet, enabling direct prediction in the ambient latent space.
  \item We demonstrate that PDG-FM improves view consistency and geometric smoothness, supported by geometric and perceptual analyses, and achieves competitive performance on standard NVS benchmarks.
\end{itemize}

\section{Related Work}
\label{sec:related_work}

\subsection{Flow Matching and Data-to-Data Generative Modeling}
Conditional Flow Matching (CFM)~\cite{lipman2023flow,liu2023flow,albergo2023building} learns continuous-time dynamics between distributions and enables flexible conditional interpolants~\cite{gagneux2025avisualdive,lipman2024flowmatchingguidecode}.
CFM has been applied to diverse domains, including general flow matching frameworks~\cite{Schusterbauer2024} and latent bridge matching~\cite{chadebec2025lbmlatentbridgematching}.
Riemannian Flow Matching (RFM)~\cite{chen2024flowmatchinggeneralgeometries} extends flow matching to curved manifolds but assumes a fixed geometry independent of the data distribution. Subsequent works introduce data-dependent metrics via metric learning~\cite{kapusniak2024metricflowmatchingsmooth} or density-based formulations~\cite{yu2025probabilitydensitygeodesicsimage,sorrenson2025learningdistancesdatanormalizing}, where geometry is defined by the underlying data density~\cite{NIPS2003densitybasedDistance,groisman2019}. 

Data-to-data generative models, including stochastic interpolants~\cite{albergo2023building} and bridge models~\cite{de2021schrodinger,zhou2023ddbm,wang2025ladb}, learn transport directly between distributions but do not impose explicit geometric constraints on the interpolation trajectories.

In contrast, we inject geometric priors without learning a Riemannian metric from scratch (e.g.,~\cite{kapusniak2024metricflowmatchingsmooth}) and enforce viewpoint-specific consistency through paired-view supervision and density-aligned trajectories.

\subsection{Novel View Synthesis with Generative Model}

The emergence of large-scale 2D generative diffusion models has led to impressive progress in generating realistic objects and scenes~\cite{rombach2022high, podell2023sdxlimprovinglatentdiffusion}. 
This evolution has been further refined by developments that improve representation quality and synthesis versatility~\cite{xu2024headrouter, xu2025context, xu2026tag, zhang2023tdec, zheng2024deep, du2026unsupervised}.
Several works~\cite{melas2023realfusion,poole2022dreamfusiontextto3dusing2d} attempt to distill priors from text-to-image diffusion models for 3D generation, but these approaches face challenges due to the semantic ambiguity of text and the computational expense of iterative optimization.

More recent efforts focus on diffusion models trained on multi-view posed image datasets.
Zero-1-to-3~\cite{liu2023zero1to3zeroshotimage3d} first demonstrated view synthesis from paired 2D views rendered from large-scale 3D datasets~\cite{deitke2023objaverse,downs2022googlescannedobjectshighquality}, yet it suffers from limited geometric consistency across viewpoints.
Subsequent works~\cite{zheng2024free3dconsistentnovelview,xiong2025light,kong2024eschernetgenerativemodelscalable,li2025nvcomposer} improve consistency and coherence via architectural and conditioning designs, yet still lack explicit modeling of cross-view geometry. Recent approaches also explore explicit 3D geometry~\cite{liu2024syncdreamergeneratingmultiviewconsistentimages,chen2025mess,fu2026dav}.

We address this gap with a probability-density-based geodesic formulation that encourages latent transitions to follow data-adaptive paths. By leveraging the underlying probability landscape to define geometry-consistent interpolants within the generative process, our formulation improves structural coherence and cross-view consistency.

\section{Background}
\label{sec:background}

\subsection{Conditional Flow Matching}

Given a coupling distribution $p_{0,1}$ of source and target data with marginals $p_0$ and $p_1$, we can sample data pair $(x_0,x_1) \sim p_{0,1}$, and specify a stochastic interpolant $x_t \sim p(x_t|t, x_0,x_1)$ at each timestep $t \in [0,1]$. 
The conditional path must satisfy the boundary conditions that $p(\cdot|t=0, x_0,x_1) = p_0$ and $p(\cdot|t=1, x_0,x_1) = p_1$.

Let $u_t(x_t|x_0,x_1)$ denote the conditional velocity field that realizes the time-dependent marginal $p(x_t|t, x_0,x_1)$, such that their equivalence is characterized by the continuity equation under standard regularity assumptions~\cite{lipman2023flow,liu2023flow}.
A neural vector field $v_\theta(x_t,t)$ is trained to approximate $u_t$ by minimizing the conditional flow matching loss:
\begin{align} 
    \ell^\text{CFM}(\theta) = \mathbb{E}_{\substack{
      x_0,x_1,t
    }}[\|v_\theta(x_t,t)-u_t(x_t|x_0,x_1)\|^2 | x_t]. 
\end{align}

\subsection{Geodesic along Data-driven Manifold}

\paragraph{Geodesic measurements} A geodesic defines the shortest path between two points on a Riemannian manifold, under a constant-speed parameterization. Such paths can be obtained by minimizing the length functional of a curve using the calculus of variations~\cite{arvanitidis2021latentspaceodditycurvature,pmlr-v151-arvanitidis22b}.

We adopt a probability-density-based geodesic~\cite{yu2025probabilitydensitygeodesicsimage,sorrenson2025learningdistancesdatanormalizing}, where the local metric tensor is defined inversely proportional to the data density. The length of a path $\gamma : [0, 1] \rightarrow \mathbb{R}^d$ is given by:
\begin{align}
S[\gamma] &= \int_0^1 L(t, \gamma(t), \gd(t)) \dt, \label{eq:geodesic}
\end{align}
where $L$ is the Lagrangian given by
\begin{align}
L(t, \gamma, \gd) &= \|\gd\|_{G(\gamma)},
\end{align}
with the norm $\|x\|_M = \sqrt{ x\transpose M x}$ scaled by a metric tensor $G(x) = p(x)^{-2} I$ inversely proportional to the density function $p : \mathbb{R}^d \rightarrow \mathbb{R}_+$.

\paragraph{Functional derivative}

The Euler--Lagrange equation that characterizes the shortest path (geodesic) for \eqref{eq:geodesic} is:
\begin{align}
\ddot{\gamma} + \|\gd\|^2\left( I - \gdh\gdh\transpose \right) \nabla \log p(\gamma) = 0,
\label{eq:euler-lagrange}
\end{align}
with $\gdh = \gd / \|\gd\|$ the unit velocity.
This second-order ODE describes the optimal path given initial conditions (position and velocity). By calculus of variations, the functional derivative $\ddelta{S}{\gamma}$ \cite{yu2025probabilitydensitygeodesicsimage} can be derived
for a constant speed parameterization as: 
\begin{align}
\ddelta{S}{\gamma} &= \frac{-1}{p(\gamma)\|\gd\|} \left( \left( I - \gdh\gdh\transpose \right) \nabla \log p(\gamma) + \frac{\gdd}{\|\gd\|^2} \right),
\label{eq:FuncDeriv}
\end{align}
which will be leveraged in Alg.~\ref{alg:geodesicNet} for training the Geodesic Network $\phi_{\eta}$.

\section{Methodology}
\label{sec:method}

\begin{figure*}[t]\centering
    \includegraphics[width=0.97\linewidth]{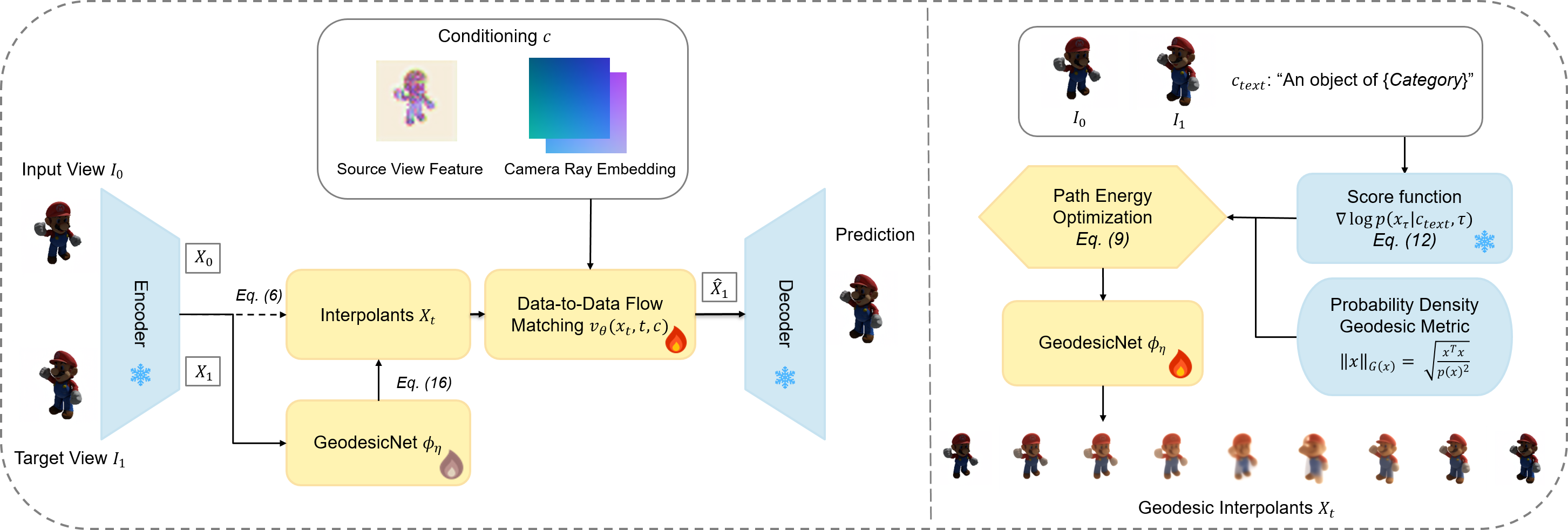}\vfill
\caption{
\textbf{Overview of the Probability Density Geodesic Flow Matching (PDG-FM) framework.} 
(a) \textit{Data-to-Data Flow Matching} framework learns deterministic mappings between paired samples $(x_0, x_1)$ by encoding source and target views into latent space and predicting intermediate states $x_t$ through either linear interpolants (Eq.~\ref{eq:flowMatching}) or geodesic interpolants (Eq.~\ref{eq:geodesic_xt}) predicted by the GeodesicNet $\phi_\eta$. 
The velocity field $v_\theta(x_t, t, c)$ is conditioned on source view features and Plücker ray embeddings, and its decoded outputs yield novel views consistent with the target pose. 
(b) \textit{Variational Distillation of Geodesics} trains GeodesicNet $\phi_\eta$ to produce manifold-aligned interpolants by minimizing path energy defined under the probability density geodesic metric 
$\|x\|_{\mathcal{G}(x)} = \sqrt{x^\top x / p(x)^2}$, 
where the density $p(x)$ is estimated using a pretrained diffusion score function (Eq.~\ref{eq:nabla_logp}).
(c) The unified PDG-FM integrates both components, resulting in geometry-aware and manifold-consistent transformations for view-consistent novel view synthesis.
}
\label{fig:framwork}
\vspace{-.2cm}
\end{figure*}

Our approach, Probability Density Geodesic Flow Matching (PDG-FM), builds upon a Data-to-Data Flow Matching framework to learn deterministic mappings between paired views. It consists of: 
(1) a base formulation with linear interpolants (Linear-D2D-FM), and
(2) Variational Distillation of Geodesics, which introduces geometry-aware interpolants via a teacher-student framework. The teacher performs geodesic optimization using a pretrained diffusion score in the DDIM-F latent space, while the student distills this into GeodesicNet in the ambient latent space. The resulting interpolants are used for flow matching, yielding view-consistent transformations for novel view synthesis.
An overview is illustrated in Figure~\ref{fig:framwork}.

\subsection{Data-to-Data Flow Matching}
\label{subsec:data2dataFM}

We adopt a Data-to-Data Flow Matching framework to learn deterministic mappings between paired views $(x_0, x_1)$ of the same scene under different camera poses, removing the need for a noise prior while explicitly enforcing structural correspondence.

\paragraph{Network Architecture}
Our velocity net $v_\theta(x_t, t, q, c)$ adopts a U-Net~\cite{ronneberger2015unetconvolutionalnetworksbiomedical} backbone similar to Zero-1-to-3. 
Each training sample consists of image pairs $(I_0, I_1)$ with corresponding camera poses $(Q_0, Q_1)$.
The model takes as input the intermediate state $x_t$ and time $t$, and is conditioned on:

\begin{itemize}
  \item $\text{PluckerRayEmbedder}(Q_0, Q_1)$: 
  Target camera ray is embedded using Plücker coordinates $r_{uv} = \psi(o, d_{uv}) = (o \times d_{uv},\, d_{uv}) \in \mathbb{R}^6$~\cite{sitzmann2021light, xiong2025light}, 
  where $o$ denotes the origin and $d_{uv}$ denotes the ray direction derived from camera intrinsics $K$ and extrinsics $(R, T)$.  
  The source view serves as the canonical reference, and target rays are expressed relative to the source camera coordinate frame.
  \item $\mathcal{E}_{\text{clip}}(I_0)$: CLIP-encoded semantic conditioning from the source view, which is concatenated with the relative pose between $(Q_0, Q_1)$ and injected through cross-attention layers to guide appearance and content alignment. 
  \item $\mathcal{E}_\text{img}(I_0)$: VAE-encoded latent from source view, which is concatenated with the interpolated latent $x_t$ as the U-Net input to preserve spatial structure.
\end{itemize}

\paragraph{Training Objective}
We first instantiate the linear interpolants baseline (Linear-D2D-FM). The conditional path and its corresponding target velocity field are defined as:
\begin{align}
    & x_t = (1-t)x_0+tx_1 + \sigma_{\text{min}}\varepsilon, \quad\varepsilon\sim \mathcal{N}(0,I), \label{eq:flowMatching}  \\
    & u_t(x_t|x_0,x_1) = x_1-x_0, 
\end{align}
with a minimal variance $\sigma_{\text{min}}=0.01$ to smooth the data samples.
In line with~\cite{ho2021cascadeddiffusionmodelshigh,Schusterbauer2024}, we augment the source latent $x_0$ according to the cosine schedule.

The velocity network $v_\theta$ is trained to regress the time derivative of the linear interpolant:
\begin{align}
\mathcal{L}_{\text{CFM}}(\theta) = \mathbb{E}_{x_0,x_1,t} \left[ \| v_\theta(x_t, t, q, c) - (x_1 - x_0) \|^2 \right]
\end{align}

\begin{algorithm}[t]
\caption{GeodesicNet Distillation}
\label{alg:geodesicNet}
\KwIn{ 
source-target Image pairs $({I}_{0,k}, {I}_{1,k})_{k\in\cK}$;
text prompts $({T}_k)_{k\in\cK}$;
DDIM timestep $\tau$; 
pretrained diffusion scaling $\beta$;
VAE encoder $\mathcal{E}_\text{img}$; 
CLIP encoder $\mathcal{E}_{\text{clip}}$.
}
\KwOut{Teacher $\phi_{\xi}$, GeodesicNet $\phi_{\eta}$.}
 \ForAll{$k \in \cK$}{
  $\{c_0, c_1\} \gets \{ \text{TextInversion}(\mathcal{E}_{\text{clip}}({T}_k), {I}_{i,k})\}_{i=\{0,1\}} $ \\
  $\{x_{0}, x_{1} \}\gets \{\mathcal{E}_\text{img}({I}_i)\}_{i=\{0,1\}}$ \\
  $\{z_{0}, z_{1} \}\gets \{\text{DDIM-F}(x_i, c_i, \tau)\}_{i=\{0,1\}}$ \\
 $ \mathcal{T} \gets \text{TimeSampler}(i)$ \\
  \ForAll{$t \in \mathcal{T}$}{
  $z_t \gets  (1 - t) z_0 + t z_1 + \phi_{\xi}(z_0, z_1, t)$ \\
  $c_t \gets (1 - t) c_0 + t c_1$ \\
  $s_t \gets \text{Score}(z_t, c_t,\tau, \beta)$  \quad{// \cref{eq:nabla_logp}} \\
  $g_t \gets \text{FuncDeriv}(z_t, v_t, a_t, s_t) $   \quad{// \cref{eq:FuncDeriv}} \\
  $\ell^\tau_t(\xi) \gets \text{StopGrad}(g)\cdot z_t$
   }
   $\ell^\tau(\xi) = \mathbb{E}_{t}[\ell^\tau_t(\xi)] $ \\
   Update $\xi$ using $\nabla_{\xi}\ell^\tau(\xi)$   \\
  \ForAll{$t \in \mathcal{T}$}{
  $x_t \gets (1 - t) x_0 + t x_1 + \phi_{\eta}(x_0, x_1, t)$ \\
  $\ell^0_t(\eta) \gets \|x_t - \text{DDIM-B}(z_t,c_t,\tau)\|^2$ \\        
  }
  $\ell^0(\eta) = \mathbb{E}_{t}[\ell^0_t(\eta)] $ \\
  Update $\eta$ using $\nabla_{\eta}\ell^0(\eta)$
 }
\end{algorithm}

\subsection{Variational Distillation of Geodesics}
\label{subsec:GeodesicOptimization}

To obtain manifold-aligned paths, we introduce GeodesicNet, a neural module that estimates probability-density geodesics using a pretrained diffusion score as a proxy for data density.
Recent work has shown the benefits of specialized interpolants for improving generation quality~\cite{bartosh2024neural}.

\paragraph{Geodesic Interpolants}
We parameterize the probability density geodesic as
\begin{align}
  x_t = (1 - t) x_0 + t x_1 + \phi_\eta(x_0, x_1, t),
\end{align}
with an ambient-space correction network $\phi_\eta$ satisfying the boundary constraints $\phi_\eta(x_0, x_1, 0)=\phi_\eta(x_0, x_1, 1)=0$. The network $\phi_\eta$ is distilled from a companion teacher network $\phi_\xi$ of the same structure but imposed in the latent space resulting from forward DDIM (DDIM-F) \cite{song2021denoising}.
This dual-network architecture separates the geometric optimization (teacher) from the efficient path generation (student), enabling robust manifold-aware interpolation.

\paragraph{Variational Distillation} We adopt a teacher-student distillation scheme for probability density geodesic optimization.
The teacher network $\phi_\xi$ can be updated in accordance with the shortest-path principle \eqref{eq:geodesic}.

Given the denoising diffusion network $\zeta$ \cite{rombach2022high},
we define the DDIM-F operator \cite{song2021denoising}, written $\text{DDIM-F}(x,c,\tau)$ with initial state $x$, condition variable $c$ and timestep $\tau$, via the first-order ODE:
\begin{align}
  x_0 &= x; \qquad \forall 0<t<\tau: \\
  \dd{}{t} \left( \frac{x_{t}}{\sqrt{\overline{\alpha}_{t}}} \right) &= \dd{}{t} \left(\sqrt{\frac{1-\overline{\alpha}_{t}}{\overline{\alpha}_{t}}} \right) \zeta(x_{t}, c, t). 
  \label{eq:ddim_inversion} 
\end{align}
The solution $x_\tau=\text{DDIM-F}(x,c,\tau)$, often seen as a ``smoothed" version of $x$ (denoted by $z$ in Alg.~\ref{alg:geodesicNet}), 
has a Stein score that can be approximated via classifier-free guidance:
\begin{align}
  \nabla \log p(x_\tau | c, \tau) &\approx \beta \omega(\tau)( \zeta(x_\tau, c, \tau) - \zeta(x_\tau, c_{\text{neg}}, \tau) ).
  \label{eq:nabla_logp} 
\end{align}
Here $c_{\text{neg}}$ is a negative prompt embedding that specifies unwanted, out-of-distribution instructions for the diffusion model.
Our choice of hyperparameters in experiments is $\beta=1, \tau=0.6$, and $\omega(\tau) = - (1-\bar{\alpha}_\tau)^{-1/2}$ a weighting function.
We choose non-zero timestep $\tau=0.6$ because score functions are poorly estimated in low-density regions~\cite{song2019generative}. By operating at moderate noise levels, we populate sparse regions of the latent space, providing more reliable gradient estimates for geodesic optimization.

This enables us to perform gradient descent update using the (rescaled) functional derivative in \eqref{eq:FuncDeriv}:
\begin{align}
  z_t &= (1 - t) z_0 + t z_1 + \phi_\xi(z_0, z_1, t), \\
  g_t &= \left( I - \gdh_t\gdh_t\transpose \right) \nabla \log p(z_t | c_t, \tau) + \frac{\gdd_t}{\|\gd_t\|^2}, \\
  \ell^\tau(\xi) &= \mathbb{E}_t[\text{StopGrad}(g_t) \cdot z_t] \to\min. \label{eq:update-xi}
\end{align}

\noindent
The alignment between teacher network $\phi_\xi$ and student network $\phi_\eta$ can be achieved via minimizing an MSE loss involving the backward DDIM (DDIM-B) operator:
\begin{align}
  x_t &= (1 - t) x_0 + t x_1 + \phi_{\eta}(x_0, x_1, t), \\
  \ell^0(\eta) &= \mathbb{E}_t[\|x_t - \text{DDIM-B}(z_t,c,\tau)\|^2] \to\min.
\label{eq:geodesic_xt}
\end{align}
More details on variational distillation are given in Alg.~\ref{alg:geodesicNet}.

\begin{algorithm}[t]
\caption{Geodesic Flow Matching}
\label{alg:geodesicFM}
\KwIn{ 
source-target image pairs $({I}_{0,k}, {I}_{1,k})_{k\in\cK}$;
source-target pose pairs $({Q}_{0,k}, {Q}_{1,k})_{k\in\cK}$;
VAE encoder $\mathcal{E}_\text{img}$; 
CLIP encoder $\mathcal{E}_{\text{clip}}$;
GeodesicNet $\phi_{\eta}$.
}
\KwOut{VelocityNet $v_{\theta}$. }
 \For{$k\in \mathcal{K}$}{
  $q_k\gets \text{PluckerRayEmbeder}({Q}_{0,k}, {Q}_{1,k})$ \\
  $c_k \gets  \mathcal{E}_{\text{clip}}( {I}_{0,k})   $ \\
  $\{x_{0,k}, x_{1,k} \}\gets \{\mathcal{E}_\text{img}({I}_{i,k})\}_{i=\{0,1\}}$ 
 }
  \While{training}{
    Sample $x_0,x_1,q,c,t$ \\
    $x_t \gets (1 - t) x_0 + t x_1 + \phi_{\eta}(x_0, x_1, t)$ \\
    $v_{\text{target}} \gets x_1 - x_0 + \nabla_{t}\phi_{\eta}(x_0, x_1, t) $ \\
    $\ell^v(\theta) \gets \mathbb{E}[ \|v_{\theta}(x_t, t, q, c) - v_{\text{target}}\|^2 ]$ \\
    Update $\theta$ using $\nabla_{\theta}\ell^v(\theta)$
  }
\end{algorithm}

\subsection{Geodesic Flow Matching for NVS}
\label{subsec:GeodesicFM}

Once the GeodesicNet $\phi_\eta$ is ready, we apply conditional flow matching \cite{liu2023flow,chen2024flowmatchinggeneralgeometries} for training the VelocityNet $v_{\theta}(x_t, t, c)$; see Alg.~\ref{alg:geodesicFM}. 
VelocityNet applies the same architecture as in Section~\ref{subsec:data2dataFM}.
The target velocity $v_{\text{target}}$ involves computation of time derivative of the network $\phi_\eta$, which can be done via forward mode of auto-differentiation (e.g.,~\texttt{jacfwd} in PyTorch).

In contrast to metric flow matching (MFM) \cite{kapusniak2024metricflowmatchingsmooth}, we adopt a two-phase approach here, that is, (1) Learn the geodesic path parameterized with an ambient-space corrector $\phi_\eta$; (2) The conditional flow matching fits $v_\theta$ to the $\phi_\eta$ guided paths. Such a design choice is more favorable in terms of: (1) It requires relatively less data and compute to train $\phi_\eta$; (2) Training and deployment of flow model $v_\theta$ is now detached from the score-dependent Riemannian metric, and hence much more efficient.

\section{Experiment}

\begin{table*}[t]
  \renewcommand{\tabcolsep}{4pt}
  \centering
  \begin{tabular}{l | l c c c c | r c c c c}
  \toprule
  & \multicolumn{5}{|c|}{Objaverse} & \multicolumn{5}{|c|}{GSO30} \\ \cmidrule{2-6} \cmidrule{7-11}
  Method & FID $\downarrow$ & CLIP-S $\uparrow$ & SSIM $\uparrow$  & PSNR $\uparrow$ & LPIPS $\downarrow$
                    & FID $\downarrow$ & CLIP-S $\uparrow$ & SSIM $\uparrow$ & PSNR $\uparrow$ & LPIPS $\downarrow$ \\ 
  \midrule 
  Zero-1-to-3              &     5.9959 &    88.6625 &     0.8446 &    19.5850 &     0.0962
                            & 12.5796 &   90.9683   & 0.8536 & 19.3038 & 0.0638 \\
  Zero-1-to-3-XL          &     7.2006 &    87.1176 &     0.7916 &    17.1722 &     0.1505
                            & 15.7340 &  89.0756    & 0.8280 & 18.6022 & 0.0910 \\
\midrule                
  EscherNet           &     7.9058 &    87.9020 &     0.8363 &    18.7560 &     0.1061
                                                &    14.7044 &    90.1891 &     0.8595 &    \textbf{20.1754} &     0.0689 \\
 Free3D &     5.5434 &    88.9575 &     0.8537 &    20.3222 &     0.0873
               &    \textbf{12.0620} &    \textbf{91.1424} &     0.8581 &    19.5610 &     0.0666 \\
\midrule              

 Naive FM   &     5.5093 &    88.8786 &     0.8622 &    20.8246 &     0.0809
               &    15.2763 &    89.8226 &     0.8631 &    19.7273 &     0.0641 \\  
  Linear-D2D-FM   &   \textbf{5.4324} &    \textbf{88.9855} &     \textbf{0.8634} &    \textbf{20.8447} &     \textbf{0.0809} 
               &  15.0543 &    89.5713 &     \textbf{0.8647} &    19.8742 &     \textbf{0.0641}   \\ 
  \bottomrule
  \end{tabular}
  \caption{
  \textbf{Novel view synthesis performance on Objaverse and GSO datasets.}
Comparison of Linear-D2D-FM against diffusion and flow baselines on Objaverse and GSO30.
  }
  \label{tab:benchmark}
  \vspace{-.2cm}
\end{table*}

We evaluate our method under a unified Data-to-Data Flow Matching framework for novel view synthesis. In this work, we distinguish between two instantiations of this framework:
(i) Linear-D2D-FM, which uses linear interpolants and serves as the base flow matching baseline, and
(ii) PDG-FM, which replaces linear interpolants with density-aware geodesic interpolants predicted by GeodesicNet.

\subsection{Data-to-Data Flow Matching for NVS}
\label{sec:d2d_fm}
\paragraph{Datasets} 
We evaluate our models on single-view novel view synthesis (NVS) using the Objaverse dataset~\cite{deitke2023objaverse}, which contains over 772,870 3D objects with multi-view renderings. Following Zero-1-to-3~\cite{liu2023zero1to3zeroshotimage3d}, we render 12 random views per object on a white background.  
For out-of-distribution evaluation, we additionally use the Google Scanned Objects (GSO) dataset~\cite{downs2022googlescannedobjectshighquality}.  
Models are trained on Objaverse and evaluated on both Objaverse test-split and GSO30 chosen by SyncDreamer~\cite{liu2024syncdreamergeneratingmultiviewconsistentimages}.

\paragraph{Experimental Setup} 
We evaluate the effectiveness of the proposed Linear-D2D-FM formulation against conventional Noise-to-Data Flow Matching (Naive FM) and diffusion-based NVS baselines.  
This experiment tests whether deterministic data-pair coupling improves view consistency and reduces artifacts in novel view generation.
We compare the Linear-D2D-FM against Zero-1-to-3~\cite{liu2023zero1to3zeroshotimage3d}, EscherNet~\cite{kong2024eschernetgenerativemodelscalable}, and Free3D~\cite{zheng2024free3dconsistentnovelview}, as well as a Noise-to-Data FM variant implemented on the same backbone.

We report quantitative results using PSNR, SSIM~\cite{ssimMetric}, LPIPS, CLIP Similarity (CLIP-S), and Fréchet Inception Distance (FID)~\cite{heusel2017gans}.   
Higher PSNR, SSIM, and CLIP-S indicate better fidelity and semantic consistency, while lower LPIPS and FID denote improved perceptual realism.

\paragraph{Implementation Details} 
All models follow the Zero-1-to-3 architecture built upon the Stable Diffusion~\cite{rombach2022high}, where text conditioning is replaced with source-view and pose conditioning.  
The CLIP [CLS] embedding and relative camera pose are linearly projected and integrated into U-Net cross-attention layers, while the VAE latents of the source and intermediate states are concatenated at the input level to preserve spatial correspondence. We employ Plücker ray embeddings through Ray Conditional Normalization (RCN)~\cite{zheng2024free3dconsistentnovelview}, enabling the model to adapt to per-pixel camera orientations.  
Models are trained under the same optimization settings (AdamW optimizer, batch size 256, learning rate $1\times10^{-5}$, 256$\times$256 resolution) and equal training steps for a fair comparison across baselines.  
We report results using released checkpoints of Zero-1-to-3, Zero-1-to-3-XL, and EscherNet, and re-train Free3D and NaiveFM using identical training protocols as our method.

\paragraph{Results} 
Table~\ref{tab:benchmark} summarizes quantitative results on Objaverse and GSO with 100 NFE (Number of Function Evaluations).  
Linear-D2D-FM consistently outperforms Naive FM and diffusion-based baselines in both fidelity and perceptual quality.  
Improvements are especially notable in FID and LPIPS, suggesting sharper detail and reduced artifacts.

\begin{figure}[h]
\hspace*{-0.5cm}
\begin{tabular}{ccccc}
  \includegraphics[width=0.16\linewidth,trim=30 30 30 30, clip]{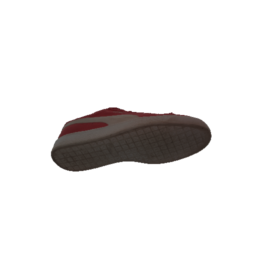} 
  & \includegraphics[width=0.16\linewidth,trim=30 30 30 30, clip]{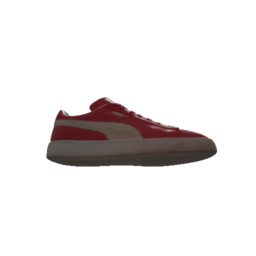} 
  &   \includegraphics[width=0.16\linewidth,trim=30 30 30 30, clip]{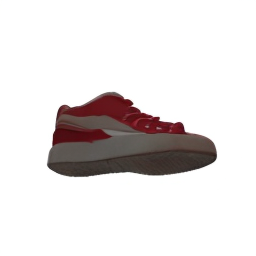} 
  &   \includegraphics[width=0.16\linewidth,trim=30 30 30 30, clip]{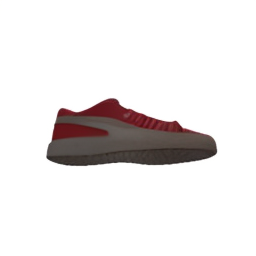} 
  &   \includegraphics[width=0.16\linewidth,trim=30 30 30 30, clip]{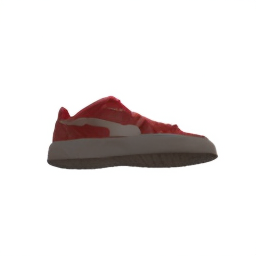} \\

      \includegraphics[width=0.16\linewidth,trim=30 30 30 30, clip]{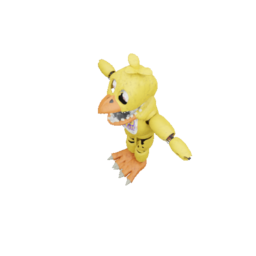} 
  & \includegraphics[width=0.16\linewidth,trim=30 30 30 30, clip]{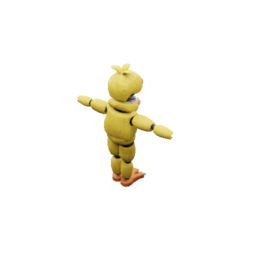} 
  &   \includegraphics[width=0.16\linewidth,trim=30 30 30 30, clip]{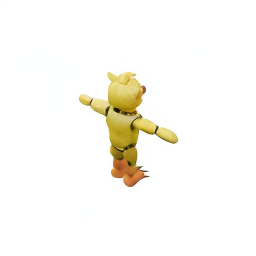} 
  &   \includegraphics[width=0.16\linewidth,trim=30 30 30 30, clip]{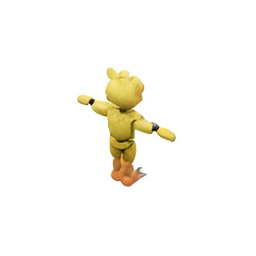} 
  &   \includegraphics[width=0.16\linewidth,trim=30 30 30 30, clip]{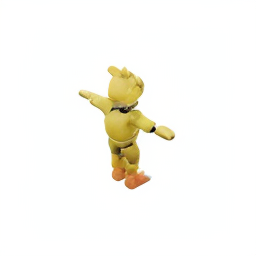} \\

 \includegraphics[width=0.16\linewidth,trim=30 30 30 30, clip]{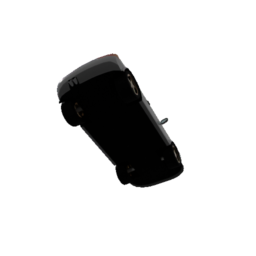} 
 & \includegraphics[width=0.16\linewidth,trim=30 30 30 30, clip]{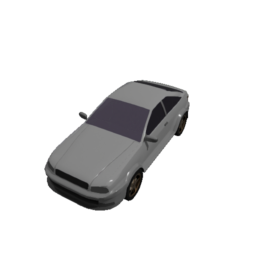} 
 &   \includegraphics[width=0.16\linewidth,trim=30 30 30 30, clip]{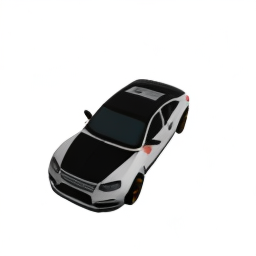} 
 &   \includegraphics[width=0.16\linewidth,trim=30 30 30 30, clip]{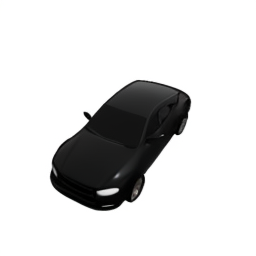}
 &   \includegraphics[width=0.16\linewidth,trim=30 30 30 30, clip]{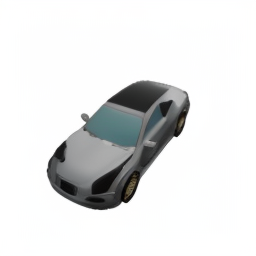} \\

  \includegraphics[width=0.16\linewidth,trim=20 20 20 20, clip]{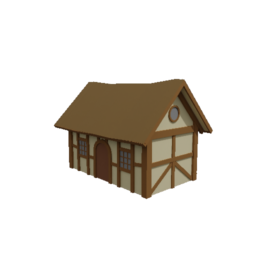} 
 & \includegraphics[width=0.16\linewidth,trim=20 20 20 20, clip]{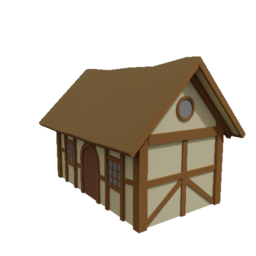} 
 &   \includegraphics[width=0.16\linewidth,trim=20 20 20 20, clip]{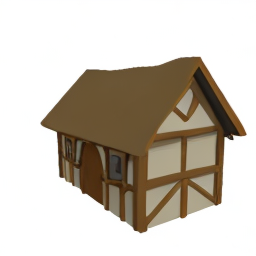} 
 &   \includegraphics[width=0.16\linewidth,trim=20 20 20 20, clip]{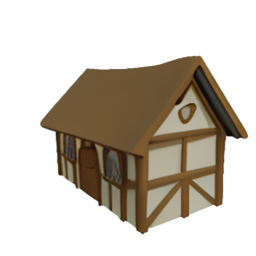}
 &   \includegraphics[width=0.16\linewidth,trim=20 20 20 20, clip]{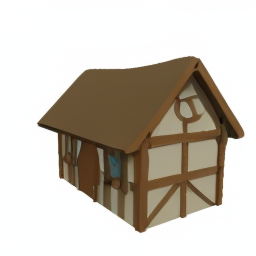} \\

    \includegraphics[width=0.16\linewidth,trim=30 30 30 30, clip]{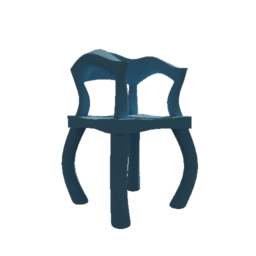} 
 & \includegraphics[width=0.16\linewidth,trim=30 30 30 30, clip]{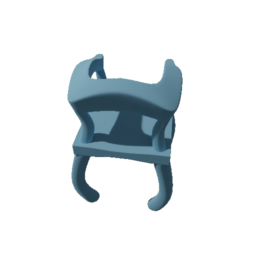} 
 &   \includegraphics[width=0.16\linewidth,trim=30 30 30 30, clip]{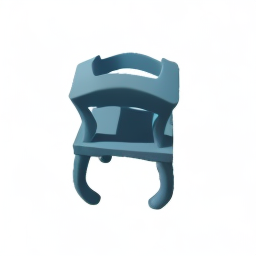} 
 &   \includegraphics[width=0.16\linewidth,trim=30 30 30 30, clip]{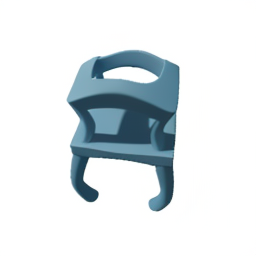}
 &   \includegraphics[width=0.16\linewidth,trim=30 30 30 30, clip]{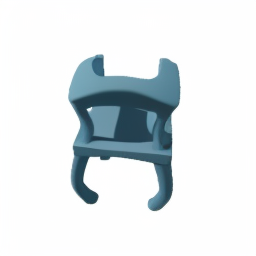} \\

Condition & Target & Free3D~\cite{zheng2024free3dconsistentnovelview} & NaiveFM & Ours \\[6pt]
\end{tabular}
\caption{
  \textbf{Qualitative comparisons} on Objaverse. 
  Comparison of Linear-D2D-FM, the Noise-to-Data FM (Naive FM) baseline and the diffusion-based Free3D model.
}
\label{fig:linear_data2data_fm}
\end{figure}

\begin{figure}
\centering
\setlength{\tabcolsep}{3pt}
\renewcommand{\arraystretch}{1.0}
\begin{tabular}{cc}
\begin{minipage}[c]{0.25\linewidth}
  \includegraphics[width=\linewidth]{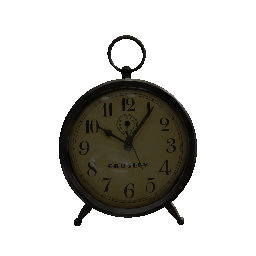}
\end{minipage} &
\begin{minipage}[c]{0.65\linewidth}
  \raisebox{0.5em}{\rotatebox{90}{\scriptsize Free3D~\cite{zheng2024free3dconsistentnovelview}}}\hfill
  \includegraphics[width=0.9\linewidth]{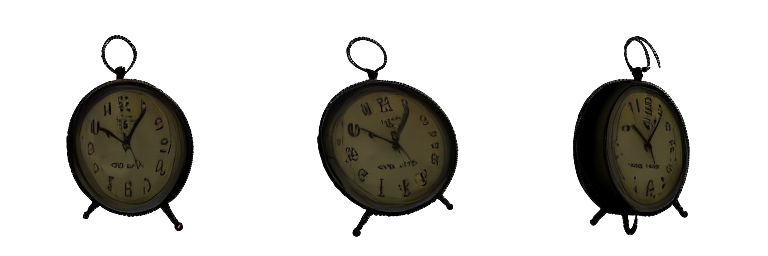}\\[2pt]
  \raisebox{0.5em}{\rotatebox{90}{\scriptsize Ours}}\hfill
  \includegraphics[width=0.9\linewidth]{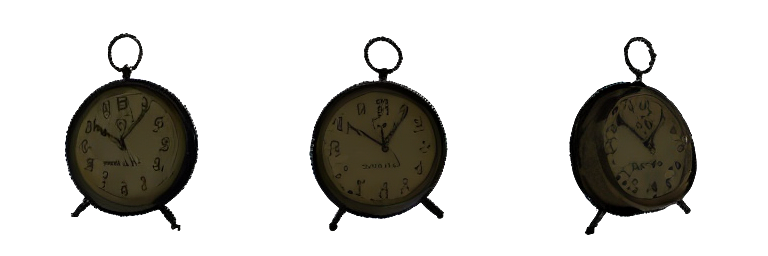}
\end{minipage}
\\[6pt]
\begin{minipage}[c]{0.25\linewidth}
  \includegraphics[width=\linewidth]{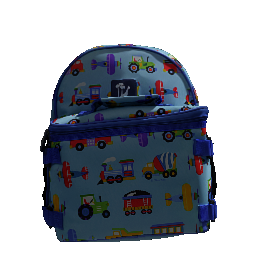}
\end{minipage} &
\begin{minipage}[c]{0.65\linewidth}
  \raisebox{0.5em}{\rotatebox{90}{\scriptsize Free3D~\cite{zheng2024free3dconsistentnovelview}}}\hfill
  \includegraphics[width=0.9\linewidth]{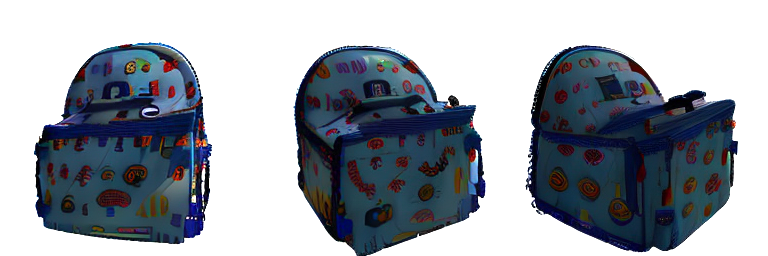}\\[2pt]
  \raisebox{0.5em}{\rotatebox{90}{\scriptsize Ours}}\hfill
  \includegraphics[width=0.9\linewidth]{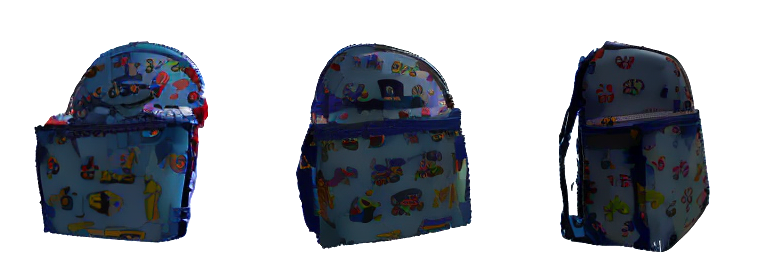}
\end{minipage}
\\[6pt]
\addlinespace[10pt]
Input View & Novel Views \\
\end{tabular}
\caption{
\textbf{Qualitative comparisons} on GSO30.
Comparison of Free3D and our method (“Ours”) given the same input view.
}
\label{fig:linear_data2data_fm_gso30}
\end{figure}

Fig.~\ref{fig:linear_data2data_fm} shows that Linear-D2D-FM better preserves structure and viewpoint alignment than diffusion methods.  
On the GSO30 dataset, our model achieves higher fidelity in novel view synthesis compared to Free3D, particularly under large viewpoint changes.  
As shown in Fig.~\ref{fig:linear_data2data_fm_gso30}, our generated views maintain better alignment with target poses and exhibit fewer artifacts, highlighting the advantage of learning deterministic flows between structured view pairs.

This advantage is more pronounced under accelerated inference. Table~\ref{tab:cfm_10nfe} shows that at 10 NFE, Linear-D2D-FM outperforms Naive FM and conditional diffusion.
These findings suggest that the data-to-data framework is a more stable and efficient formulation for the NVS task, providing a solid foundation for further geodesic interpolants design.

\begin{table}
  \renewcommand{\tabcolsep}{4pt}
  \centering
  \begin{tabular}{l | c c c c}
  \toprule
  & \multicolumn{4}{|c|}{Objaverse} \\ \cmidrule{2-5}
  Method & FID $\downarrow$ & CLIP-S $\uparrow$ & SSIM $\uparrow$ & LPIPS $\downarrow$ \\
  \midrule
  Free3D  &  22.4533   &  77.9569     & 0.4470     &  0.4372  \\
  Free3D  &  8.9367   &  84.4197     & 0.7894     &  0.1493  \\
  Naive FM  &  5.7768 & 88.9998 & 0.8679 & 0.0779 \\
  Linear-D2D-FM & 5.8223 & 88.9749 & 0.8688 & 0.0782  \\
  \bottomrule     
  \end{tabular}
  \caption{
      \textbf{Novel view synthesis performance with 10 NFE.}
  Linear-D2D-FM preserves superior performance with 10 NFE.
  }
  \label{tab:cfm_10nfe}
\vspace{-.2cm}
\end{table}

\subsection{Geodesic Data-to-Data Flow Matching}
\label{sec:pdg_d2d_fm}

\paragraph{Experimental Setup}
We conduct experiments on the LVIS-annotated Objaverse subset to assess the effect of incorporating probability density-based geodesic interpolants into the data-to-data framework.
We employ a pretrained text-to-image Stable Diffusion model~\cite{rombach2022high} for both the VAE encoder-decoder and the score function proxy for the underlying data density.

\paragraph{Implementation Details}
Following Algorithm~\ref{alg:geodesicNet}, we train a pair of networks, $\phi_{\xi}$ and $\phi_{\eta}$ jointly.
The network $\phi_{\xi}$ predicts geodesic interpolants $z_t$ in the diffusion latent space by minimizing the Euler-Lagrange residual~\ref{eq:FuncDeriv}, while $\phi_{\eta}$ maps these latent trajectories to the corresponding VAE space $x_t$ and their time derivatives $v_t$.
Condition embeddings are obtained via text inversion~\cite{galimage} of CLIP-encoded prompts of the form “an object of \texttt{{category}}”, ensuring consistent conditioning across all object views.

Once the GeodesicNet $\phi_{\eta}$ is trained, we use its predicted geodesic interpolants to train \textbf{PDG-FM}, as shown in Algorithm~\ref{alg:geodesicFM}, and we compare it with \textbf{Linear-D2D-FM} whose interpolants are computed by linear latent interpolation. 
For \textbf{Metric FM} (which learns the metric from scratch)~\cite{kapusniak2024metricflowmatchingsmooth}, we follow the original protocol adapted to paired translation: an RBF metric is first trained without OT coupling using $\kappa=0.5$, $\epsilon=10^{-4}$, and k-means clustering; geodesic interpolants are then parameterized with a neural network; finally, flow matching is trained on these interpolants, mirroring the PDG-FM pipeline.
All flow matching models are initialized from the pretrained velocity network used in the base Linear-D2D-FM experiment.

\paragraph{Results}
Table~\ref{tab:geodesic_FM} reports the quantitative results comparing PDG-FM with its linear counterpart.
Incorporating geodesic interpolants yields consistent improvements in CLIP similarity, SSIM, and PSNR, demonstrating enhanced perceptual quality and geometric coherence in novel view generation.
These gains suggest that accounting for the curvature of the underlying probability density guides the flow model toward more semantically consistent and physically plausible trajectories.
The benefit of geodesic alignment is pronounced in scenarios requiring precise structural transitions, while linear paths are easier to approximate in simple cases, resulting in smaller gains from geodesic alignment.
Moreover, the improved consistency can be traced to the interpolation geometry analyzed in Section~\ref{sec:geodesic_analysis}, where geodesic paths were shown to exhibit higher optical flow magnitude and lower Euler-Lagrange residuals than linear interpolants.
Together, these findings indicate that the learned probability density geodesics provide an effective prior for regularizing data-to-data transformations in flow matching, leading to novel views that remain faithful to input geometry while maintaining visual realism.

\begin{figure}
\begin{tabular}{cccc}
    \includegraphics[width=0.2\linewidth]{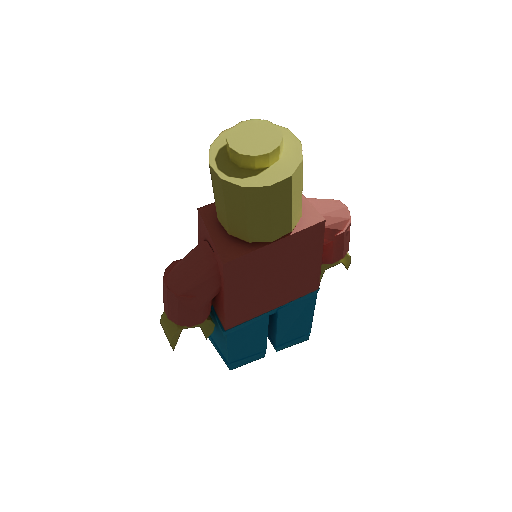} 
  & \includegraphics[width=0.2\linewidth]{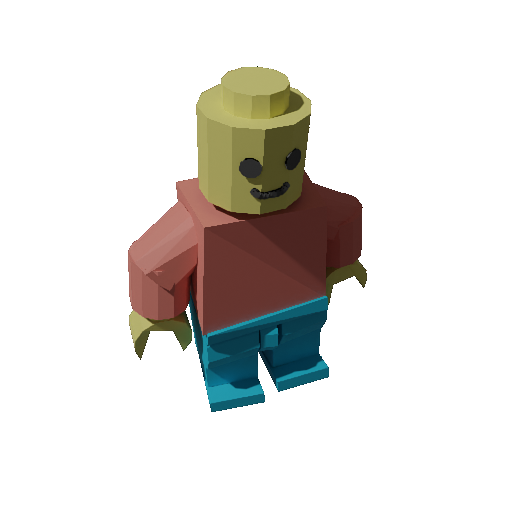} 
  &   \includegraphics[width=0.2\linewidth]{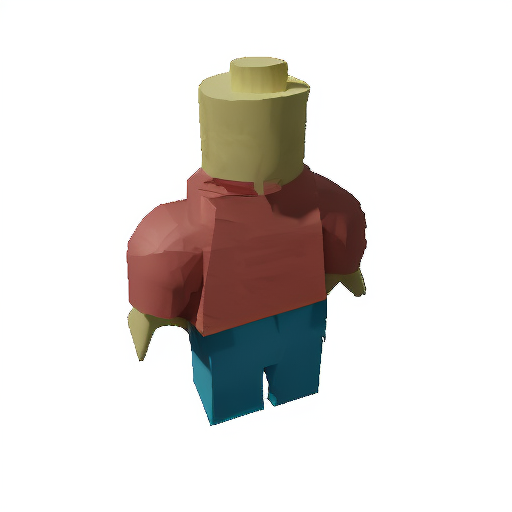}
  &   \includegraphics[width=0.2\linewidth]{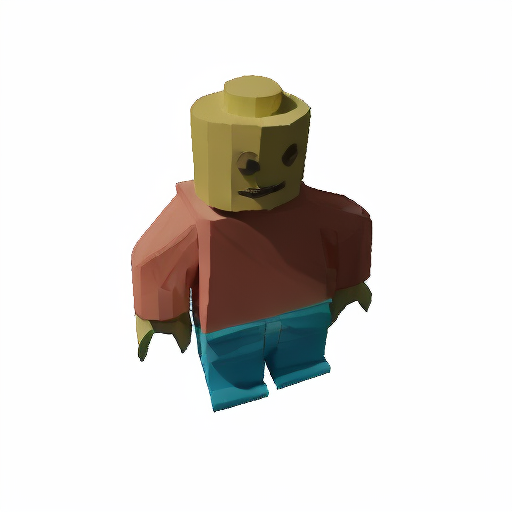}  
   \\

 \includegraphics[width=0.2\linewidth]{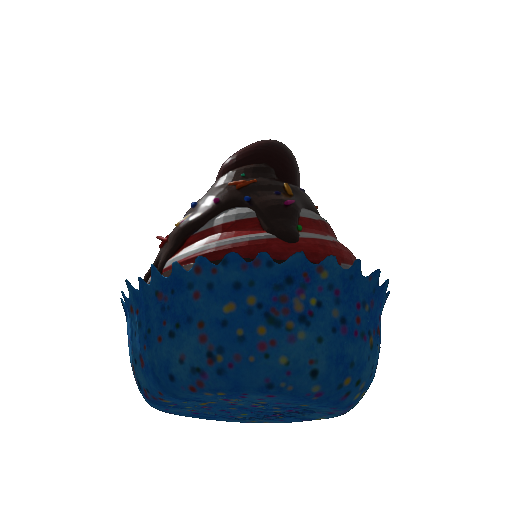} 
 & \includegraphics[width=0.2\linewidth]{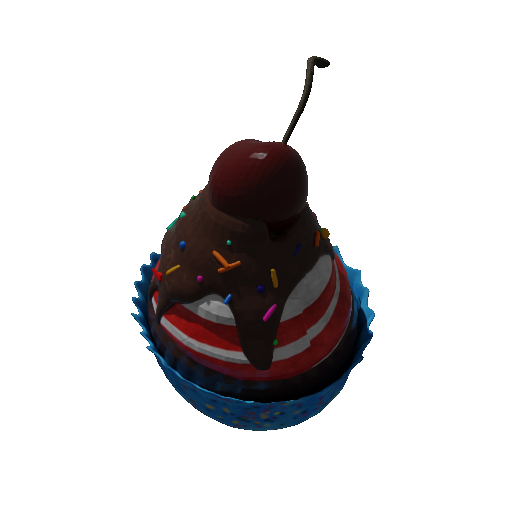} 
 &   \includegraphics[width=0.2\linewidth]{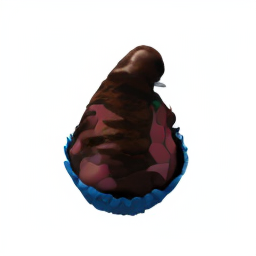}
 &   \includegraphics[width=0.2\linewidth]{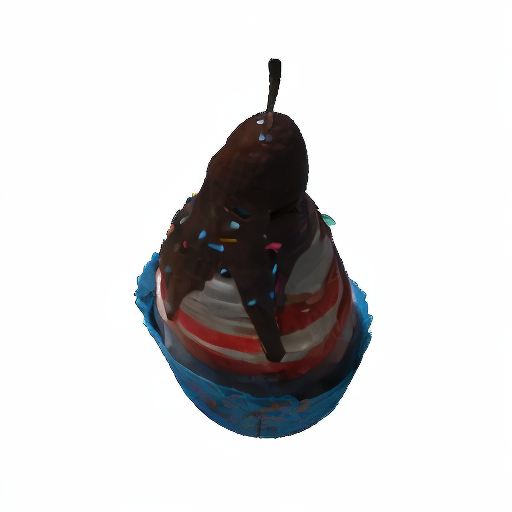} 
\\

  \includegraphics[width=0.2\linewidth,trim=30 30 30 30, clip]{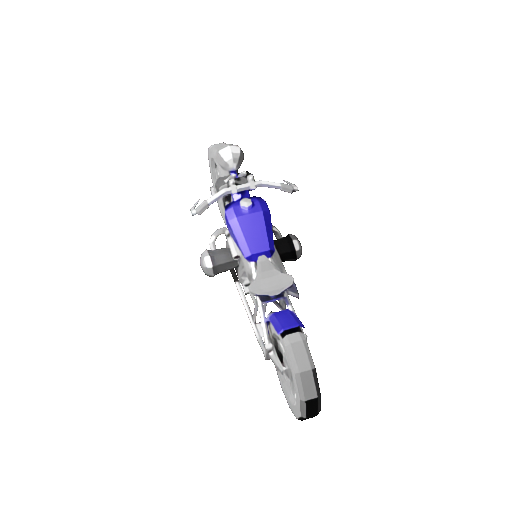} 
 & \includegraphics[width=0.2\linewidth,trim=30 30 30 30, clip]{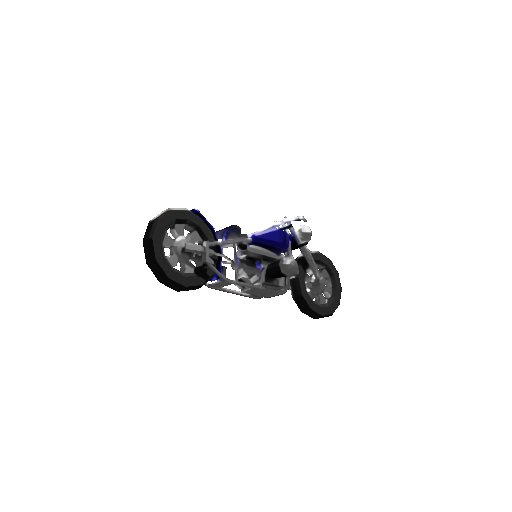} 
 &   \includegraphics[width=0.2\linewidth,trim=30 30 30 30, clip]{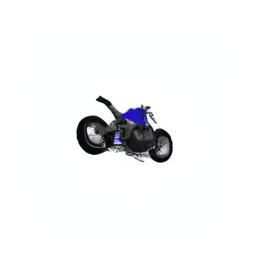}
 &   \includegraphics[width=0.2\linewidth,trim=30 30 30 30, clip]{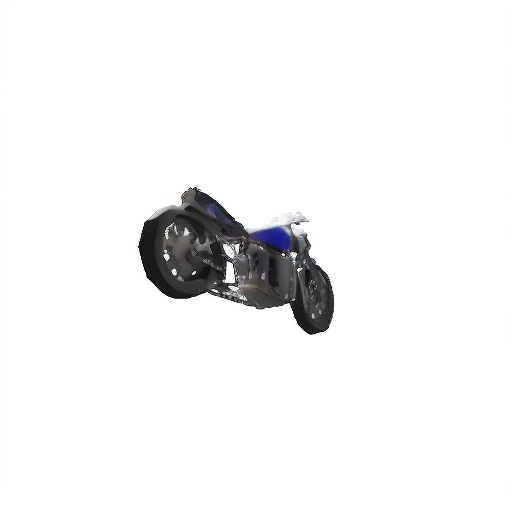} \\

   \includegraphics[width=0.2\linewidth,trim=30 30 30 30, clip]{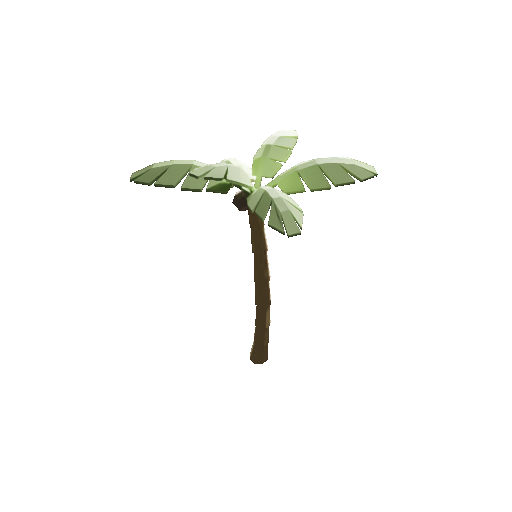} 
 & \includegraphics[width=0.2\linewidth,trim=30 30 30 30, clip]{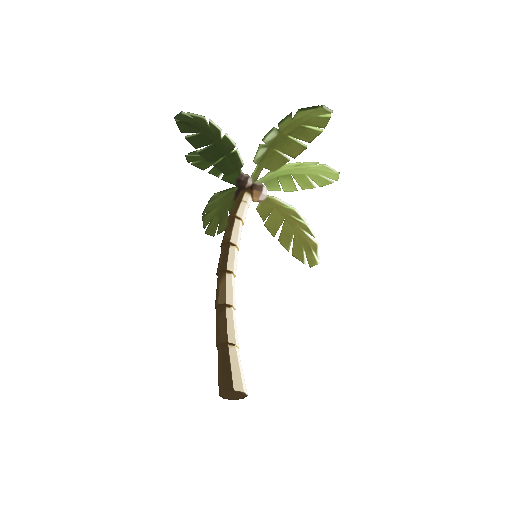} 
 &   \includegraphics[width=0.2\linewidth,trim=30 30 30 30, clip]{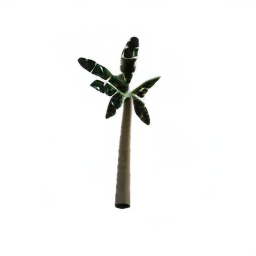}
 &   \includegraphics[width=0.2\linewidth,trim=30 30 30 30, clip]{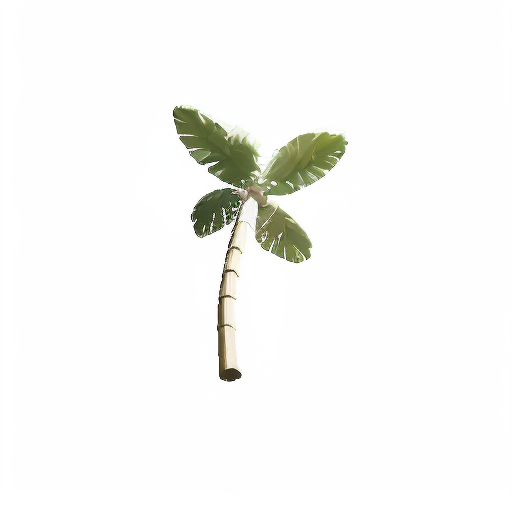} \\

Condition & Target & Linear & Geodesic  \\[6pt]
\end{tabular}
\caption{
\textbf{Qualitative comparisons} between \textbf{PDG-FM (geodesic)} and Linear-D2D-FM (linear) on Objaverse.  
Visual results showing that PDG-FM generates more geometrically faithful novel views.
The improvement reflects the effect of energy-guided optimization along data-dependent geodesics.
}
\label{fig:geodesic_improvement}
\end{figure}

\begin{table}[t]
  \renewcommand{\tabcolsep}{4pt}
  \centering
  \begin{tabular}{l | c c c c}
  \toprule
  & \multicolumn{4}{|c|}{Objaverse}  \\ \cmidrule{2-5}
  Method & FID $\downarrow$ & CLIP-S $\uparrow$ & SSIM $\uparrow$ & LPIPS $\downarrow$ \\ 
  \midrule 
    Linear-D2D-FM & 11.8124 & 94.3476 & 0.8736 & 0.0809 \\
    Metric FM & 11.9609 & 92.0370 & 0.8663 & 0.0900 \\
    PDG-FM & 10.4010 & 92.3368 & 0.8768 & 0.0804 \\
  \bottomrule     
  \end{tabular}
  \caption{
  \textbf{Effect of probability density geodesic alignment (PDG-FM) compared to linear interpolants (Linear-D2D-FM) and Metric FM under a reduced training setup.}  
  }
  \label{tab:geodesic_FM}
\end{table}

\begin{figure*}[th!]\centering
\setlength{\tabcolsep}{6pt}

\begin{tabular}{>{\centering\arraybackslash}m{0.05\linewidth} m{0.85\linewidth}}
    \rotatebox{90}{Linear} &
    \begin{minipage}{\linewidth}
        \includegraphics[width=\linewidth]{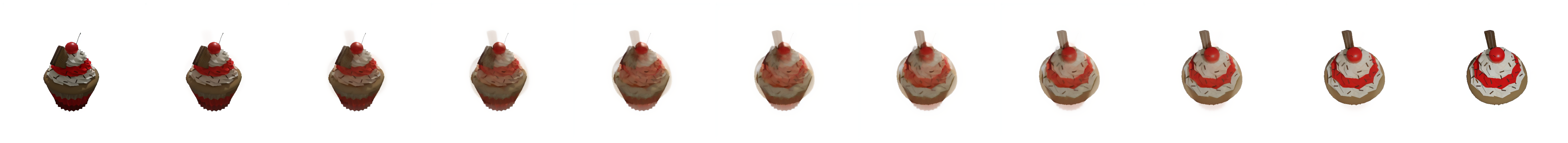}
        \includegraphics[width=\linewidth]{assets/path_figure/mario_linear_007to010.png}
    \end{minipage}
    \\[2em]

    \rotatebox{90}{Geodesic} &
    \begin{minipage}{\linewidth}
        \includegraphics[width=\linewidth]{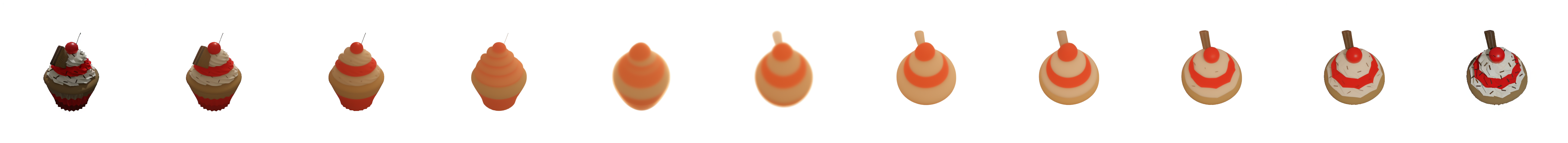}
        \includegraphics[width=\linewidth]{assets/path_figure/mario_TI_007to010.png}
    \end{minipage}
\end{tabular}
\vspace{-.4cm}
\caption{
\textbf{Geodesic Interpolants vs Linear Interpolants.}
Geodesic interpolants traverse semantically meaningful regions of the data manifold, 
producing perceptually consistent interpolants. 
In contrast, linear interpolants merely blend the two endpoints, resulting in limited structural continuity and unrealistic transitions.
}
\label{fig:interpolants_dataset}
\vspace{-.2cm}
\end{figure*}

\subsection{Probability Density Geodesic}
\label{sec:geodesic_analysis}
We study the behavior of the proposed geodesic interpolants using the training procedure described in Algorithm~\ref{alg:geodesicNet}. The analysis focuses on both the geometric and energetic properties of the learned paths across the GSO30 dataset.

\paragraph{Interpolation Path Geometry}

Perceptual Path Length (PPL)~\cite{karras2019stylebasedgeneratorarchitecturegenerative} measures interpolation smoothness, but low PPL may correspond to simple 2D cross-fading without 3D consistency. 
We posit that dynamic viewpoint change indicates a more geometrically meaningful path. To quantify this, we compute the Average Optical Flow Magnitude (AOFM) across 11 interpolated frames using a pretrained RAFT model~\cite{teed2020raft}, where higher values indicate motion consistent with camera rotation rather than static blending.

Table~\ref{tab:geodesicPath} shows that geodesic interpolants achieve higher AOFM than linear ones. 
As visualized in Fig.~\ref{fig:interpolants_dataset}, they traverse semantically meaningful latents on the data manifold and better match GT camera-induced trajectories, whereas linear interpolants mainly blend endpoints. In particular, linear interpolants exhibit low AOFM, consistent with artifacts, while geodesic ones closely align with GT trajectories, reflecting the intended geometry-aware behavior.

\begin{table}[h]
  \renewcommand{\tabcolsep}{4pt}
  \centering
  \begin{tabular}{l c | c c}
  \toprule
  Method & Setting & PPL $\downarrow$  & AOFM $\uparrow$ \\ 
  \midrule
   Ground-Truth  & 6 objs  & 0.2575 & 14.4716 \\     
   Linear            & 6 objs  & 0.2130 & 1.0411 \\     
   Initial    & 6 objs & 0.5710 & 6.4773 \\
   Geodesic        & 6 objs & 0.5023 & 13.6968 \\  

  \bottomrule     
  \end{tabular}
  \caption{
  \textbf{Geodesic Path Study Results.}
  We report PPL and AOFM for linear, DDIM-initialized, and geodesic interpolants, with ground-truth camera trajectories as the reference.
  }
  \label{tab:geodesicPath}
\end{table}

\paragraph{Analysis of Path Energy}
We further analyze the geodesic gradient norm, and it is defined as the magnitude of the functional derivative in Eq.~\ref{eq:FuncDeriv}. 
A lower norm signifies better satisfaction of the Euler-Lagrange condition and thus closer adherence to the high-density regions.

\begin{figure}[h!]\centering
  \vspace{-.4cm}
    \includegraphics[width=0.8\linewidth]{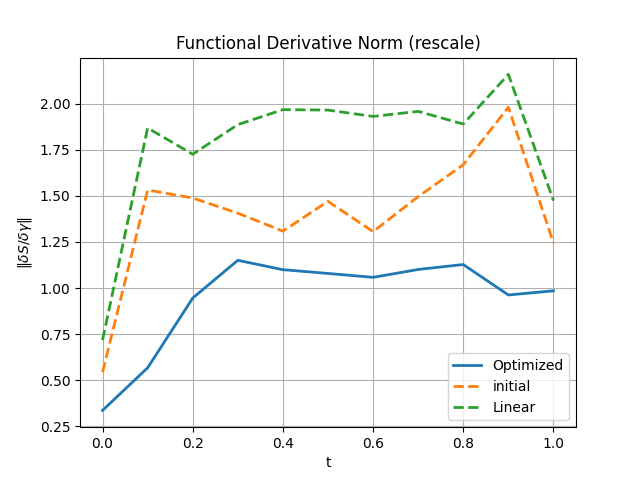}\vfill
  \vspace{-.2cm}
    \caption{
    \textbf{Comparison of Geodesic Gradient Norm across Time}
    for geodesic interpolants, their pre-optimization counterparts, and linear interpolants on training dataset. 
    The optimized geodesic paths have lower residuals thus indicating better satisfaction of the Euler-Lagrange condition, aligning with improved perceptual and geometric consistency.
    }
    \label{fig:geodesic_gradient_norm}
  \vspace{-.4cm}
\end{figure}

As shown in Fig.~\ref{fig:geodesic_gradient_norm}, geodesic interpolants maintain lower energy residuals than both pre-optimization paths and linear interpolants, confirming that the optimization in Alg.~\ref{alg:geodesicNet} effectively regularizes the interpolation trajectory.

\section{Conclusion}
We presented a framework for studying latent-space structure in generative models via geodesic-informed flow matching. By leveraging a pretrained score function as a proxy for data density, our method allows the exploration of meaningful paths in the latent space, demonstrated on a viewpoint transformation task. Experiments show that geodesic-informed interpolation can capture semantic structure more faithfully than naïve linear flows, highlighting the potential of latent-space geometry for guiding generative processes.

While the current approach involves multiple training stages and remains computationally intensive, scaling density-based geodesic estimation to larger datasets and improving efficiency under low-step inference remain important directions for future work, paving the way toward more efficient formulations of geodesic-informed generative modeling.

{
    \small
    \bibliographystyle{ieeenat_fullname}
    \bibliography{main}
}

\clearpage
\setcounter{page}{1}
\maketitlesupplementary


\section{Derivations}

This section provides detailed derivations of the Euler--Lagrange equation and functional derivative for the density-based geodesic formulation.

We recall the weighted length functional
\begin{align}
    S[\gamma] &= \int_0^1 L\big(t,\gamma(t),\dot\gamma(t)\big)\,dt, 
    \qquad
    \gamma:[0,1]\to\mathbb{R}^d
\end{align}
with the Lagrangian written in three equivalent forms:
\begin{align}
    L(t,\gamma,\dot\gamma)
    &= \|\dot\gamma\|_{G(\gamma)} \\
    &= p(\gamma)^{-1}\,\|\dot\gamma\| \label{eq:L-pform}\\
    &= \sqrt{\dot\gamma^\top G(\gamma)\,\dot\gamma},
\end{align}
where the metric tensor is defined as
\begin{align}
    G(\gamma) = p(\gamma)^{-2}\,I.
\end{align}

\subsection{Euler-Lagrange Equation}

The Euler-Lagrange equation is given by:
\begin{equation}
    \frac{d}{dt}\left(\frac{\partial L}{\partial \dot\gamma}\right) - \frac{\partial L}{\partial \gamma} = 0.
\end{equation}

Compute partial derivatives using $L = p(\gamma)^{-1}\|\dot\gamma\|$:
\begin{align}
    \frac{\partial L}{\partial \dot\gamma} &= p(\gamma)^{-1} \hat{\dot\gamma}, \\
    \frac{\partial L}{\partial \gamma} &= -p(\gamma)^{-1} \nabla \log p(\gamma) \|\dot\gamma\|.
\end{align}

The time derivative yields:
\begin{align}
\begin{split}
    \frac{d}{dt}\left(\frac{\partial L}{\partial \dot\gamma}\right) 
    &= -p(\gamma)^{-1}(\nabla \log p(\gamma)\cdot\dot\gamma)\hat{\dot\gamma} \\
    &\quad + p(\gamma)^{-1} \frac{(I - \hat{\dot\gamma}\hat{\dot\gamma}^\top)\ddot\gamma}{\|\dot\gamma\|}.
\end{split}
\end{align}

Substituting into Euler--Lagrange:
\begin{align}
\begin{split}
  -p(\gamma)^{-1}(\nabla \log p(\gamma)\cdot\dot\gamma)\hat{\dot\gamma} 
  &+ p(\gamma)^{-1} \frac{(I - \hat{\dot\gamma}\hat{\dot\gamma}^\top)\ddot\gamma}{\|\dot\gamma\|} \\
  & + p(\gamma)^{-1} \nabla \log p(\gamma) \|\dot\gamma\| = 0.
\end{split}
\end{align}

Multiply by $p(\gamma)$ and simplify using $\hat{\dot\gamma} = \frac{\dot\gamma}{\|\dot\gamma\|}$:
\begin{align}
    \frac{(I - \hat{\dot\gamma}\hat{\dot\gamma}^\top)\ddot\gamma}{\|\dot\gamma\|} 
    - (\nabla \log p(\gamma)\cdot\dot\gamma)\hat{\dot\gamma} + \nabla \log p(\gamma)\|\dot\gamma\| = 0.
\end{align}

Note the identity:
\begin{equation}
    \nabla \log p(\gamma)\|\dot\gamma\| - (\nabla \log p(\gamma)\cdot\dot\gamma)\hat{\dot\gamma} 
    = \|\dot\gamma\|(I - \hat{\dot\gamma}\hat{\dot\gamma}^\top)\nabla \log p(\gamma).
\end{equation}

Thus:
\begin{equation}
    \frac{(I - \hat{\dot\gamma}\hat{\dot\gamma}^\top)\ddot\gamma}{\|\dot\gamma\|} 
    + \|\dot\gamma\|(I - \hat{\dot\gamma}\hat{\dot\gamma}^\top)\nabla \log p(\gamma) = 0.
\end{equation}

Multiplying by $\|\dot\gamma\|$ and using the fact that for constant-speed paths, $(I - \hat{\dot\gamma}\hat{\dot\gamma}^\top)\ddot\gamma = \ddot\gamma$, recovers the equivalent formulas in the main text \eqref{eq:euler-lagrange}:
\begin{equation}
    \boxed{\ddot\gamma + \|\dot\gamma\|^2(I - \hat{\dot\gamma}\hat{\dot\gamma}^\top)\nabla \log p(\gamma) = 0.}
    \label{eq:euler-lagrange-derived}
\end{equation}

\subsection{Functional Derivative}

The functional derivative is:
\begin{equation}
    \frac{\delta S}{\delta \gamma} = \frac{\partial L}{\partial \gamma} - \frac{d}{dt}\left(\frac{\partial L}{\partial \dot\gamma}\right).
\end{equation}

Using our previous results:
\begin{align}
\begin{split}
    \frac{\delta S}{\delta \gamma} = -p(\gamma)^{-1} \nabla \log p(\gamma) \|\dot\gamma\| 
    &+ p(\gamma)^{-1}(\nabla \log p(\gamma)\cdot\dot\gamma)\hat{\dot\gamma} \\
    &- p(\gamma)^{-1} \frac{(I - \hat{\dot\gamma}\hat{\dot\gamma}^\top)\ddot\gamma}{\|\dot\gamma\|}.
\end{split}
\end{align}

Factor out the projection identity  $(I - \hat{\dot\gamma}\hat{\dot\gamma}^\top)$:
\begin{align}
    \frac{\delta S}{\delta \gamma} = -\frac{1}{p(\gamma)\|\dot\gamma\|} \big[ &(I - \hat{\dot\gamma}\hat{\dot\gamma}^\top)
    \left( \nabla \log p(\gamma) \|\dot\gamma\|^2 
     + \ddot\gamma \right) \big].
\end{align}

For constant-speed paths, $(I - \hat{\dot\gamma}\hat{\dot\gamma}^\top)\ddot\gamma = \ddot\gamma$, yielding:
\begin{equation}
    \boxed{ \frac{\delta S}{\delta \gamma} = \frac{-1}{p(\gamma)\|\dot\gamma\|} \left[ (I - \hat{\dot\gamma}\hat{\dot\gamma}^\top)\nabla \log p(\gamma) + \frac{\ddot\gamma}{\|\dot\gamma\|^2} \right]. }
    \label{eq:func-deriv-derived}
\end{equation}

This represents the normal component of the path variation, as noted in the main text \eqref{eq:FuncDeriv}, consistent with the geometric decomposition in the Euler--Lagrange equation.

\subsection{Probability density along the path}
To compute the functional derivative norm, we estimate the probability density along paths. Following~\cite{chao2023investigating,yu2025probabilitydensitygeodesicsimage}, we compute relative probabilities $p(\gamma(t))/p(\gamma(a))$ since absolute density estimation is challenging.

For the conservative vector field $\nabla \log p$, the relative log-probability between $\gamma(a)$ and $\gamma(t)$ is path-independent:
\begin{align}
\log \tilde{p}_a(\gamma(b)) &:=
\log p(\gamma(b)) - \log p(\gamma(a))\\
&= \int_a^b  \gd(t)\transpose \nabla \log p(\gamma(t)) \dt .
\label{eq:log_probability}
\end{align}

We approximate this integral using the trapezoidal rule. Let $f(\tau) = \dot\gamma(\tau)\transpose \nabla \log p(\gamma(\tau))$ and sample the path at $n+1$ equally spaced points $t_i = a + i/n$ for $i=0,\dots,n$. The relative log-probability at $\gamma(t_i)$ is:
\begin{align}
\log \tilde{p}_a(\gamma(t_i)) &\approx \! \frac{1}{n} \left(\! \frac{f(a)}{2}  \!+\! \sum_{k=1}^{i-1} f(t_k) \!+\! \frac{f(t_i)}{2}  \!\right) .
\end{align}

This enables computation of $\tilde{p}_a(\gamma(t_i)) = p(\gamma(t_i)) / p(\gamma(a))$ along the curve, which is used in evaluating \eqref{eq:FuncDeriv} for comparison of geodesic gradient norm in Fig.~\ref{fig:geodesic_gradient_norm}.

\section{Further Implementation Details}
\label{sec:implementation_details}

This section provides comprehensive implementation details for the key components of our Probability Density Geodesic Flow Matching (PDG-FM) framework, complementing the methodological descriptions in Sec.~\ref{sec:method}.

\subsection{Data-to-Data Flow Matching Implementation}
\label{supp:data2data_impl}

\paragraph{Training Configuration}
Our Linear-D2D-FM, as well as Free3D and Naive FM, are trained using the linear interpolant formulation from Eq.~\ref{eq:flowMatching}. All models inherit the publicly released checkpoint from Zero-1-to-3~\cite{liu2023zero1to3zeroshotimage3d} to ensure fair comparison.

Training is conducted in the latent space at $32 \times 32$ resolution, obtained by encoding $256 \times 256$ input images through the VAE from Stable Diffusion~\cite{rombach2022high}, which provides an 8$\times$ downscaling factor. We train all models for 20,000 steps with a global batch size of 256, using the AdamW optimizer with learning rate $1\times10^{-5}$ and standard weight decay.

\subsection{Variational Distillation of Geodesics}
\label{supp:geodesic_impl}

The geodesic distillation follows Algorithm~\ref{alg:geodesicNet} using the LVIS-annotated Objaverse subset with 12 random views per object and category annotations.

\paragraph{Preprocessing Pipeline}
Image pairs at $512 \times 512$ resolution are encoded to $64 \times 64$ ambient latents $x_0, x_1$ using the VAE encoder. For each object pair with shared category text prompt, we apply Text Inversion \cite{galimage} to fine-tune CLIP text embeddings, yielding dedicated embeddings $c_0, c_1$ for respective views. The inversion runs for 500 steps with learning rate 0.005 using AdamW optimizer.

The ambient latents $x_0, x_1$ and their corresponding text embeddings $c_0, c_1$ are then processed through DDIM-F($x, c, \tau$) with $\tau=0.6$, classifier-free guidance scale $\text{cfg}=1$, and 30 NFE (using a 50-step DDIM scheduler) to obtain the final latent pairs $z_0, z_1$ in the DDIM-F latent space.

\paragraph{GeodesicNet Optimization}
The teacher network $\phi_\xi$ operates in the DDIM-F latent space and is optimized using AdamW with learning rate $1\times10^{-6}$ and uniform timestep sampling. The optimization uses score gradients computed via Eq.~\ref{eq:nabla_logp}, with conditioning employing $c_t - c_{\text{neg}}$ where $c_t=(1-t)c_0+t c_1$ represents time-linear text conditioning. The negative prompt targets unrealistic image artifacts: ``{A doubling image, unrealistic, artifacts, distortions, unnatural blending, ghosting effects, overlapping edges, harsh transitions, motion blur, poor resolution, low detail}''.

The student network $\phi_\eta$ operates in ambient latent space and is optimized using AdamW with learning rate $1\times10^{-3}$, minimizing the MSE loss between predicted $x_t$ and DDIM-B($z_t,c_t,\tau$) reconstructions. Both networks use identical UNet architectures with 128 channels and 2 residual blocks.

\subsection{Probability Density Geodesic Flow Matching}
\label{supp:pdg_fm_impl}
We train two Data-to-Data Flow Matching variants: Linear-D2D-FM using linear interpolants;
and PDG-FM (geodesic) using geodesic interpolants from the trained GeodesicNet $\phi_\eta$.
Both models initialize from the Zero-1-to-3 checkpoint and train at $64 \times 64$ latent space for 40,000 steps with global batch size 32.

\subsection{Computational Efficiency}
\label{supp:inference_impl}
All models are trained on NVIDIA RTX A6000 GPUs with 48GB memory.

Our two-phase approach (geodesic distillation followed by flow matching) provides significant efficiency advantages: GeodesicNet training requires fewer samples than end-to-end metric flow matching; flow model training is detached from score function evaluation, avoiding expensive density gradient computations during FM training; and the modular design enables independent improvements to geodesic estimation and flow matching components.

\section{Further Experiment Results}

\paragraph{CFM Ablation} 
We analyze two key design choices in Data-to-Data Flow Matching: noise augmentation and training timestep sampling. Table~\ref{tab:cfm_ablation} presents the quantitative evaluation.

For noise augmentation, we compare our default setting (eps=400) with a lower-noise variant (eps=50).
Both configurations are evaluated using 100 inference steps. Results demonstrate that moderate noise augmentation (eps=400) consistently outperforms the low-noise alternative across all metrics, indicating that appropriate noise regularization prevents overfitting while preserving semantic consistency.

For timestep sampling, we compare our default lognormal sampling (std=1, mean=0) against discrete uniform sampling strategies optimized for few-step inference. The discrete distributions $U_{10}$ and $U_{4}$ sample training timesteps that align with specific inference steps (10 and 4 NFE respectively). Our default lognormal sampling is evaluated using 10 inference steps.  Notably, these discrete strategies outperform continuous uniform sampling when evaluated at their target inference steps, with $U_{10}$ achieving the best overall balance across evaluation metrics.

\begin{table}[h]
  \renewcommand{\tabcolsep}{4pt}
  \centering
  \footnotesize
  \begin{tabular}{l H | c c c c H }
  \toprule
  & & \multicolumn{5}{|c|}{Objaverse} \\ \cmidrule{3-7} 
  Method & NFE & FID $\downarrow$ & CLIP-S $\uparrow$ & SSIM $\uparrow$  & PSNR $\uparrow$ & LPIPS $\downarrow$
                  \\
  \midrule 
  Linear-D2D-FM  & 100   &  5.4324 &   88.9855 &    0.8634 &   20.8447 &  0.0809  \\ 
  Linear-D2D-FM(eps=50)     & 100 &     5.8050 &    88.3582 &     0.8532 &    20.0291 &     0.0883 \\
  \midrule
Linear-D2D-FM &  10  &   5.8223    &   88.9749   &  0.8688   &   21.3045   & 0.0782  \\
Linear-D2D-FM, $U_{10}$  &  10   & 5.5146 &    88.9185 &     0.8706 &    21.5952 &     0.0737 \\
Linear-D2D-FM, $U_{4}$  & 4   &     7.4414 &    88.1954 &     0.8784 &    22.2092 &     0.0743  \\
  \bottomrule     
  \end{tabular}
  \caption{
  \textbf{Ablation study of noise augmentation (top) and timestep sampling (bottom) in Linear-D2D-FM.}
  }
  \label{tab:cfm_ablation}
  \vspace{-.2cm}
\end{table}

\begin{figure*}[t]
\hspace*{-18mm}
\small
\begin{tabular}{cccccccc}
  \includegraphics[width=0.11\linewidth,trim=20 20 20 20, clip]{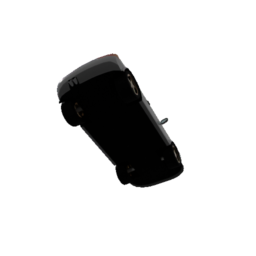} 
  & \includegraphics[width=0.11\linewidth,trim=20 20 20 20, clip]{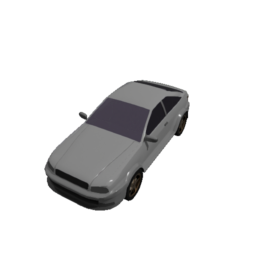} 
  &   \includegraphics[width=0.11\linewidth,trim=20 20 20 20, clip]{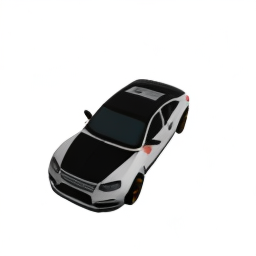} 
  &   \includegraphics[width=0.11\linewidth,trim=20 20 20 20, clip]{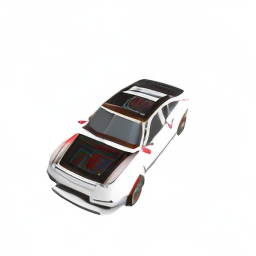}
  &   \includegraphics[width=0.11\linewidth,trim=20 20 20 20, clip]{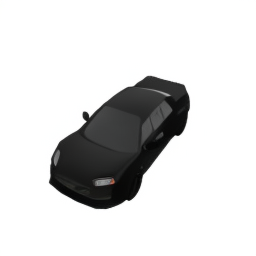} 
  &   \includegraphics[width=0.11\linewidth,trim=20 20 20 20, clip]{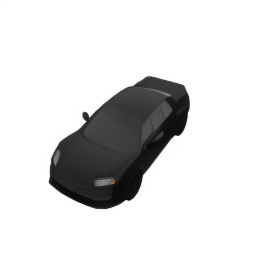} 
  &   \includegraphics[width=0.11\linewidth,trim=20 20 20 20, clip]{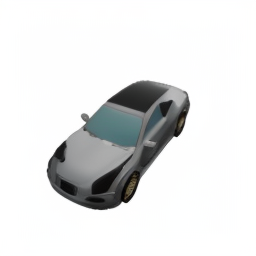}
  &   \includegraphics[width=0.11\linewidth,trim=20 20 20 20, clip]{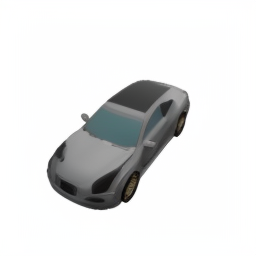} \\
 
  \includegraphics[width=0.11\linewidth,trim=30 30 30 30, clip]{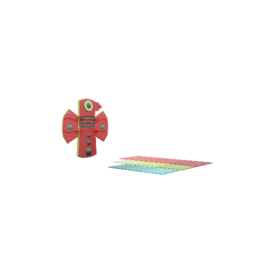} 
  & \includegraphics[width=0.11\linewidth,trim=30 30 30 30, clip]{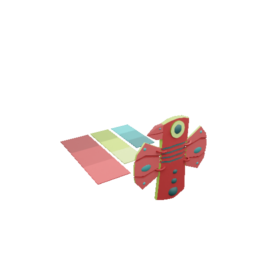} 
  &   \includegraphics[width=0.11\linewidth,trim=30 30 30 30, clip]{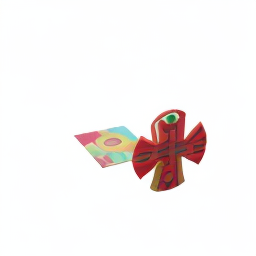} 
  &   \includegraphics[width=0.11\linewidth,trim=30 30 30 30, clip]{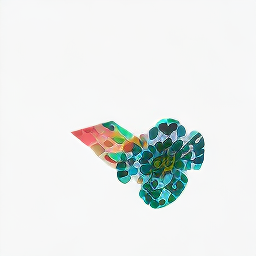}
  &   \includegraphics[width=0.11\linewidth,trim=30 30 30 30, clip]{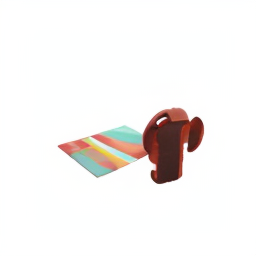} 
  &   \includegraphics[width=0.11\linewidth,trim=30 30 30 30, clip]{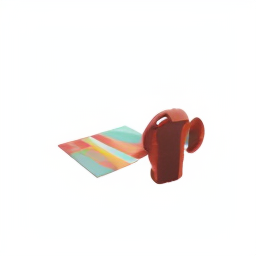} 
  &   \includegraphics[width=0.11\linewidth,trim=30 30 30 30, clip]{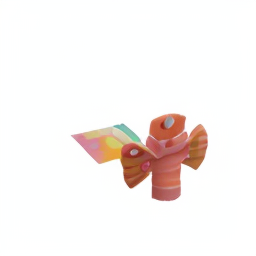}
  &   \includegraphics[width=0.11\linewidth,trim=30 30 30 30, clip]{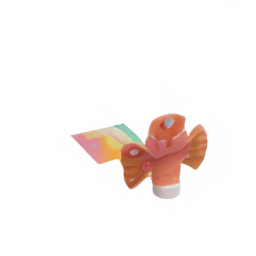} \\

   \includegraphics[width=0.11\linewidth,trim=20 20 20 20, clip]{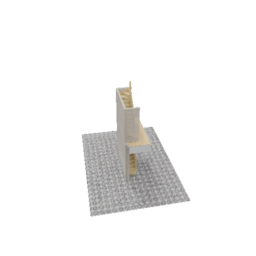} 
  & \includegraphics[width=0.11\linewidth,trim=20 20 20 20, clip]{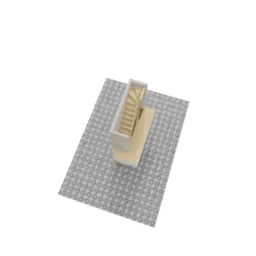} 
  &   \includegraphics[width=0.11\linewidth,trim=20 20 20 20, clip]{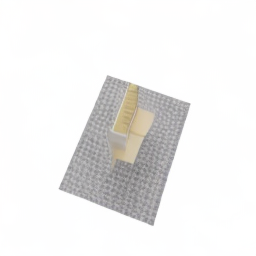} 
  &   \includegraphics[width=0.11\linewidth,trim=20 20 20 20, clip]{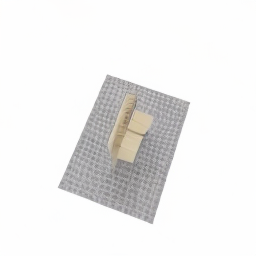}
  &   \includegraphics[width=0.11\linewidth,trim=20 20 20 20, clip]{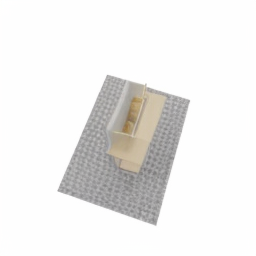} 
  &   \includegraphics[width=0.11\linewidth,trim=20 20 20 20, clip]{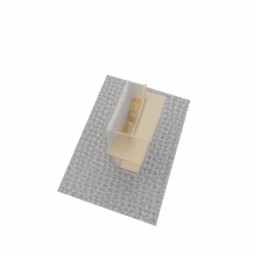} 
  &   \includegraphics[width=0.11\linewidth,trim=20 20 20 20, clip]{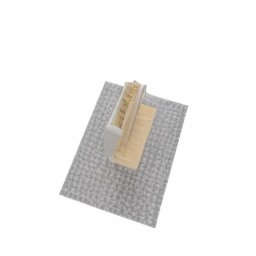}
  &   \includegraphics[width=0.11\linewidth,trim=20 20 20 20, clip]{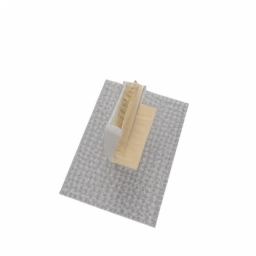} \\

  \includegraphics[width=0.11\linewidth]{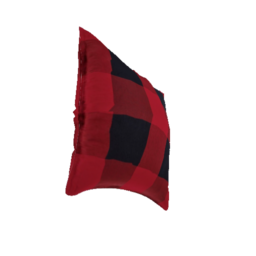} 
  & \includegraphics[width=0.11\linewidth]{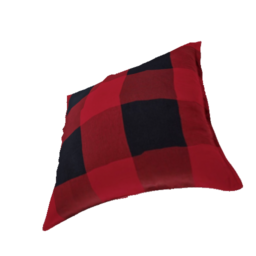} 
  &   \includegraphics[width=0.11\linewidth]{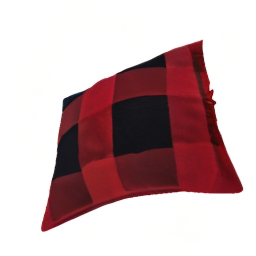} 
  &   \includegraphics[width=0.11\linewidth]{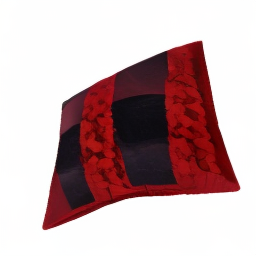}
  &   \includegraphics[width=0.11\linewidth]{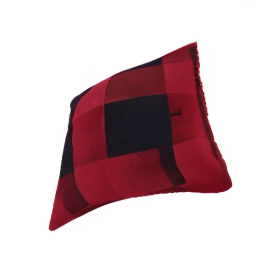} 
  &   \includegraphics[width=0.11\linewidth]{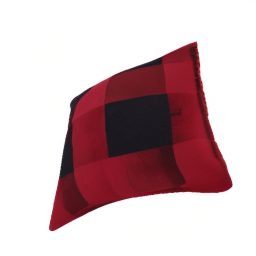} 
  &   \includegraphics[width=0.11\linewidth]{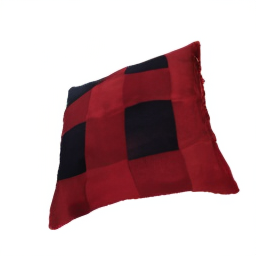}
  &   \includegraphics[width=0.11\linewidth]{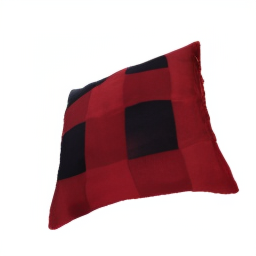} \\

    \includegraphics[width=0.11\linewidth]{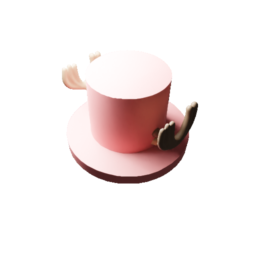} 
  & \includegraphics[width=0.11\linewidth]{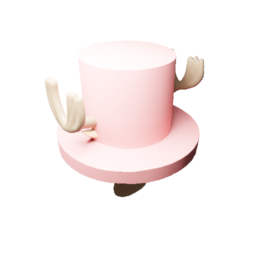} 
  &   \includegraphics[width=0.11\linewidth]{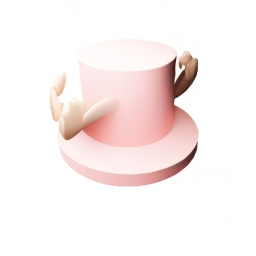} 
  &   \includegraphics[width=0.11\linewidth]{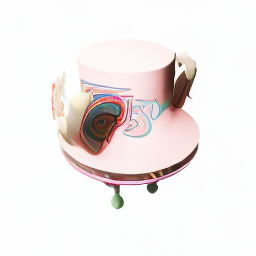}
  &   \includegraphics[width=0.11\linewidth]{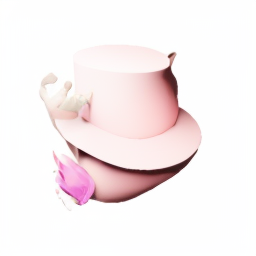} 
  &   \includegraphics[width=0.11\linewidth]{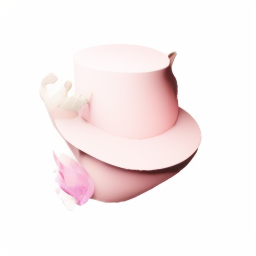} 
  &   \includegraphics[width=0.11\linewidth]{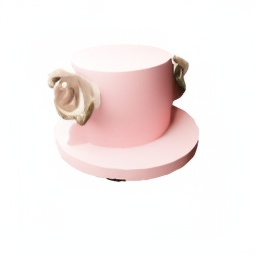}
  &   \includegraphics[width=0.11\linewidth]{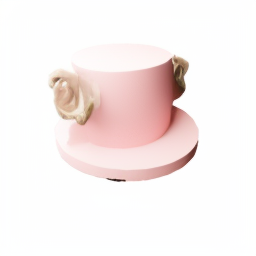} \\

  \includegraphics[width=0.11\linewidth]{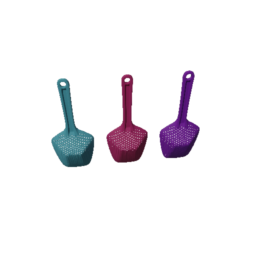} 
  & \includegraphics[width=0.11\linewidth]{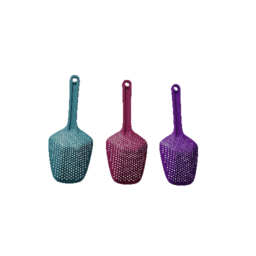} 
  &   \includegraphics[width=0.11\linewidth]{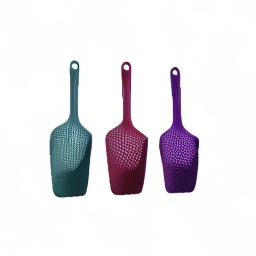} 
  &   \includegraphics[width=0.11\linewidth]{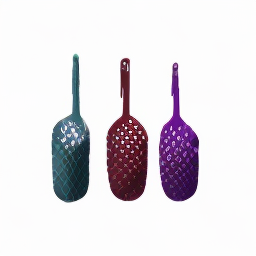}
  &   \includegraphics[width=0.11\linewidth]{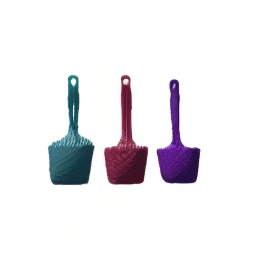} 
  &   \includegraphics[width=0.11\linewidth]{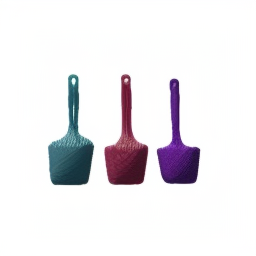} 
  &   \includegraphics[width=0.11\linewidth]{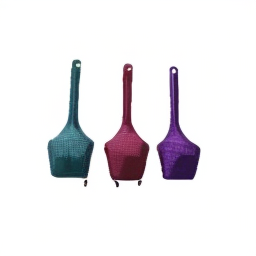}
  &   \includegraphics[width=0.11\linewidth]{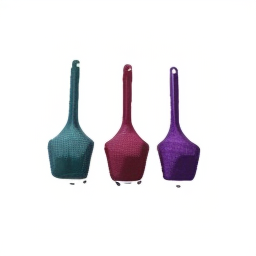} \\

Condition & Target & Free3D & Free3D (10NFE) & NaiveFM & NaiveFM (10NFE) & Linear-D2D-FM & Linear-D2D-FM (10NFE) \\
\end{tabular}
\caption{
  \textbf{Extended Qualitative Results} on Objaverse.
}
\label{fig:more_d2dfm}
\end{figure*}

\begin{figure*}[t]
\hspace*{-18mm}
\small
\begin{tabular}{cccccccc}
  \includegraphics[width=0.11\linewidth,trim=10 10 10 10, clip]{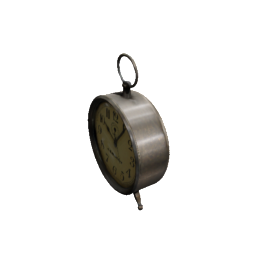} 
  & \includegraphics[width=0.11\linewidth,trim=10 10 10 10, clip]{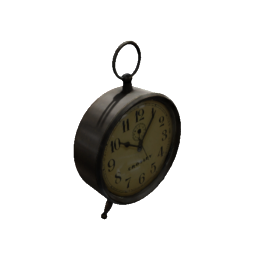} 
  &   \includegraphics[width=0.11\linewidth,trim=10 10 10 10, clip]{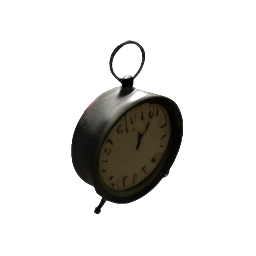} 
  &   \includegraphics[width=0.11\linewidth,trim=10 10 10 10, clip]{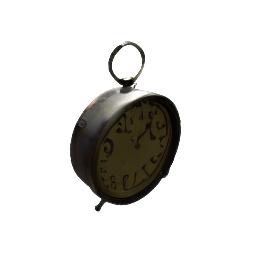}
  &   \includegraphics[width=0.11\linewidth,trim=10 10 10 10, clip]{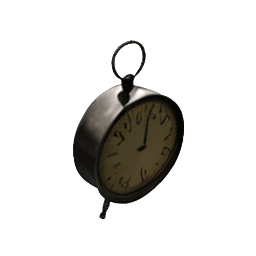} 
  &   \includegraphics[width=0.11\linewidth,trim=10 10 10 10, clip]{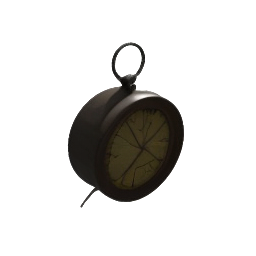} 
  &   \includegraphics[width=0.11\linewidth,trim=10 10 10 10, clip]{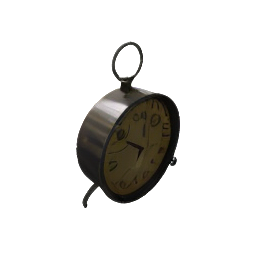}
  &   \includegraphics[width=0.11\linewidth,trim=10 10 10 10, clip]{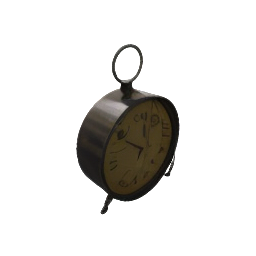} \\

    \includegraphics[width=0.11\linewidth]{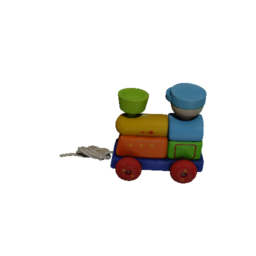} 
  & \includegraphics[width=0.11\linewidth]{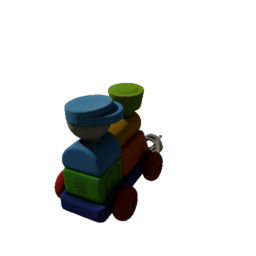} 
  &   \includegraphics[width=0.11\linewidth]{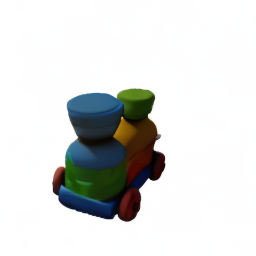} 
  &   \includegraphics[width=0.11\linewidth]{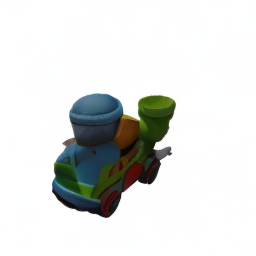}
  &   \includegraphics[width=0.11\linewidth]{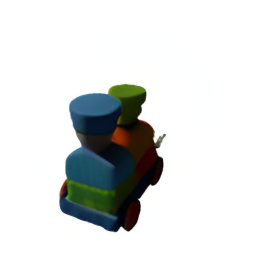} 
  &   \includegraphics[width=0.11\linewidth]{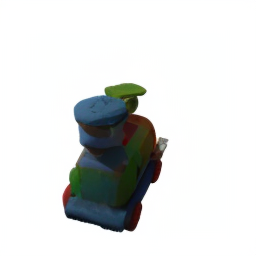} 
  &   \includegraphics[width=0.11\linewidth]{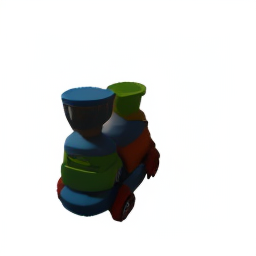}
  &   \includegraphics[width=0.11\linewidth]{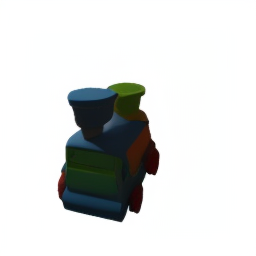} \\

      \includegraphics[width=0.11\linewidth]{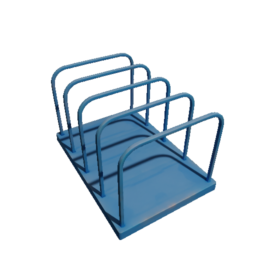} 
  & \includegraphics[width=0.11\linewidth]{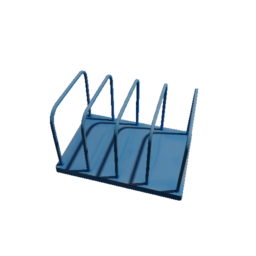} 
  &   \includegraphics[width=0.11\linewidth]{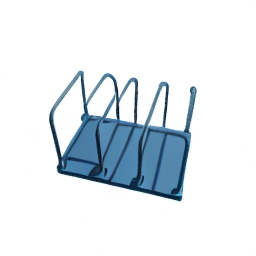} 
  &   \includegraphics[width=0.11\linewidth]{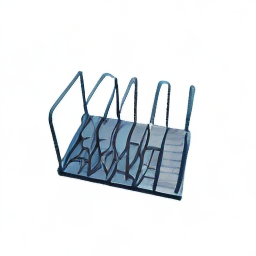}
  &   \includegraphics[width=0.11\linewidth]{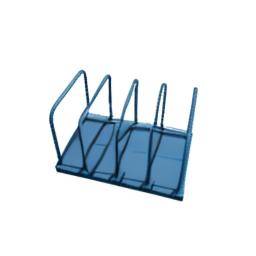} 
  &   \includegraphics[width=0.11\linewidth]{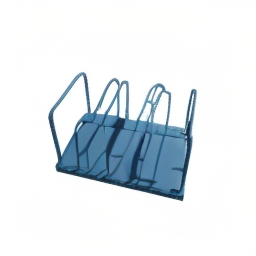} 
  &   \includegraphics[width=0.11\linewidth]{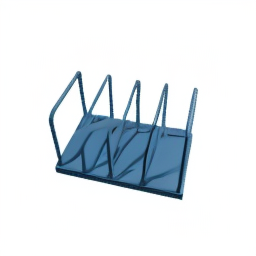}
  &   \includegraphics[width=0.11\linewidth]{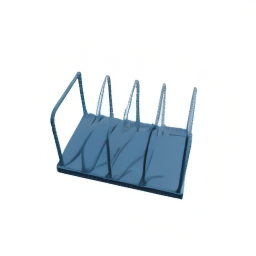} \\

Condition & Target & Free3D & Free3D (10NFE) & NaiveFM & NaiveFM (10NFE) & Linear-D2D-FM & Linear-D2D-FM (10NFE) \\
\end{tabular}
\caption{
  \textbf{Extended Qualitative Results} on GSO30.
}
\label{fig:more_d2dfm_gso30}
\end{figure*}

\paragraph{Extended Linear-D2D-FM Qualitative Results}
Figures~\ref{fig:more_d2dfm} and~\ref{fig:more_d2dfm_gso30} provide additional qualitative comparisons of our Data-to-Data FM against Noise-to-Data FM and Free3D baselines. 
Our method maintains superior structural consistency and detail preservation, particularly in the challenging 10 NFE setting. For Free3D, we use cfg=1 at 10NFE as conditional diffusion models exhibit significant performance degradation with classifier-free guidance during accelerated inference. 
The deterministic coupling in Linear-D2D-FM enables more faithful geometry reconstruction and reduces artifacts observed in diffusion-based approaches under limited inference budgets.

\begin{figure}[h!]
\vspace*{-10pt}
\begin{tabular}{cccc}
\vspace*{-15pt}
  \includegraphics[width=0.22\linewidth,trim=10 10 10 50, clip]{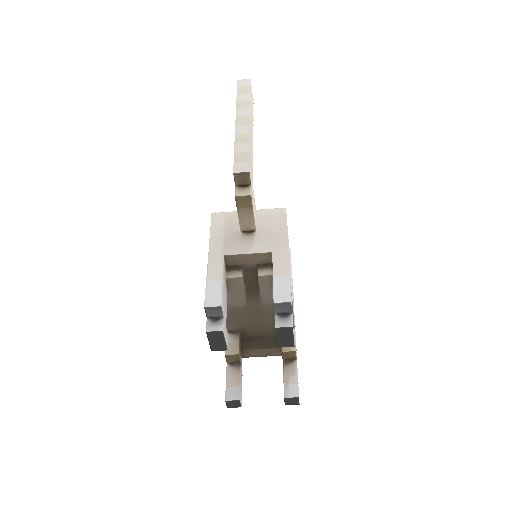} 
  & \includegraphics[width=0.22\linewidth,trim=10 10 10 50, clip]{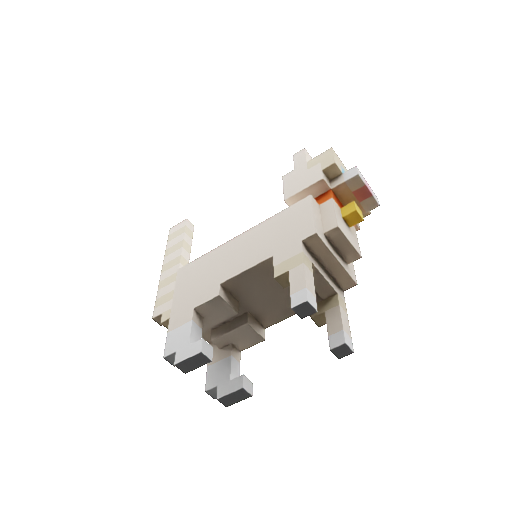} 
  &   \includegraphics[width=0.22\linewidth,trim=10 10 10 50, clip]{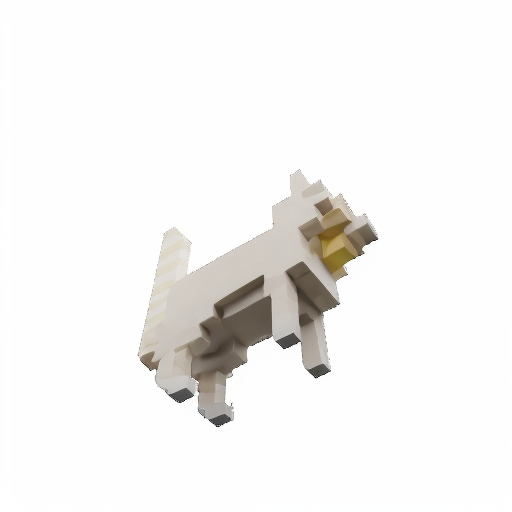} 
  &   \includegraphics[width=0.22\linewidth,trim=10 10 10 50, clip]{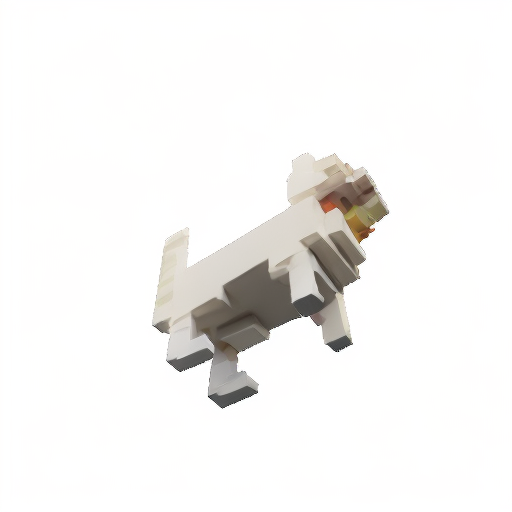}  \\  \vspace*{-15pt}

       \includegraphics[width=0.22\linewidth,trim=10 10 10 10, clip]{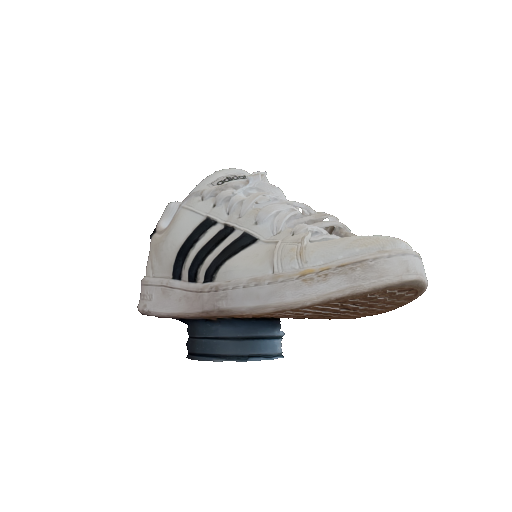} 
  & \includegraphics[width=0.22\linewidth,trim=10 10 10 10, clip]{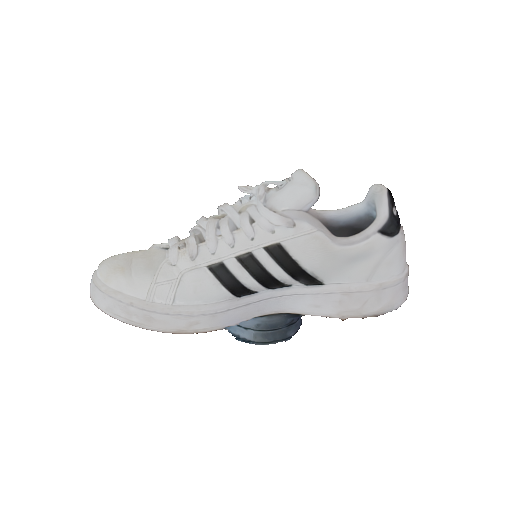} 
  &   \includegraphics[width=0.22\linewidth,trim=10 10 10 10, clip]{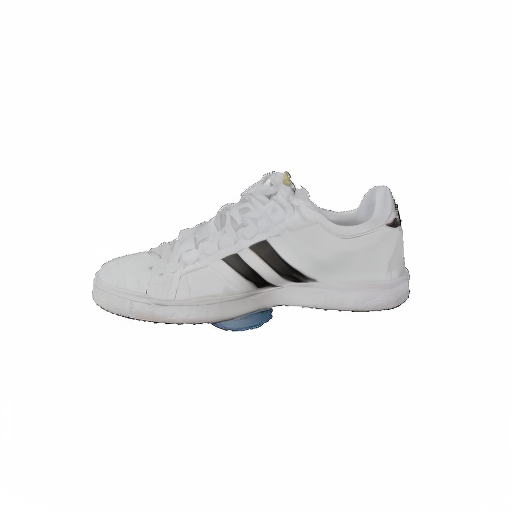} 
  &   \includegraphics[width=0.22\linewidth,trim=10 10 10 10, clip]{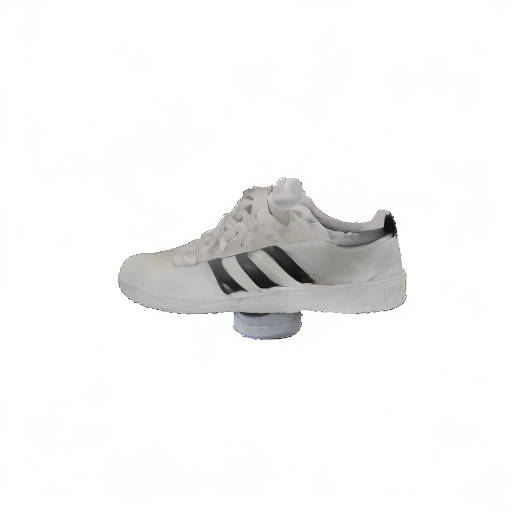} \\  \vspace*{-15pt}

 \includegraphics[width=0.22\linewidth,trim=10 10 10 10, clip]{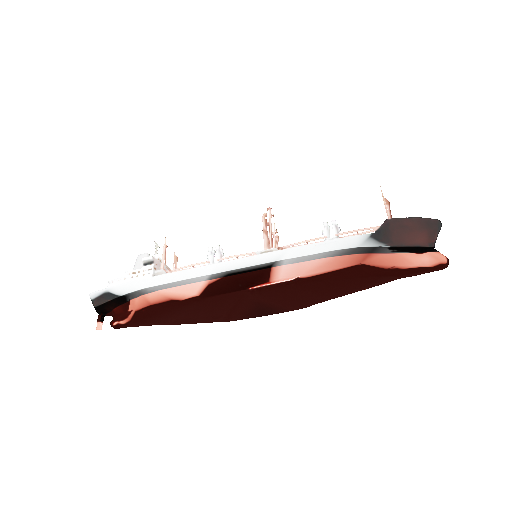} 
  & \includegraphics[width=0.22\linewidth,trim=10 10 10 10, clip]{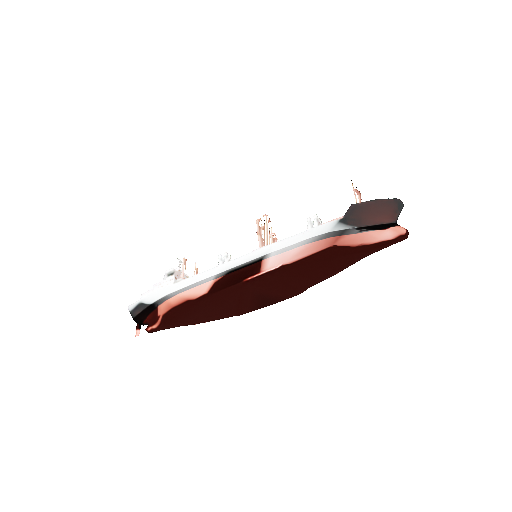} 
  &   \includegraphics[width=0.22\linewidth,trim=10 10 10 10, clip]{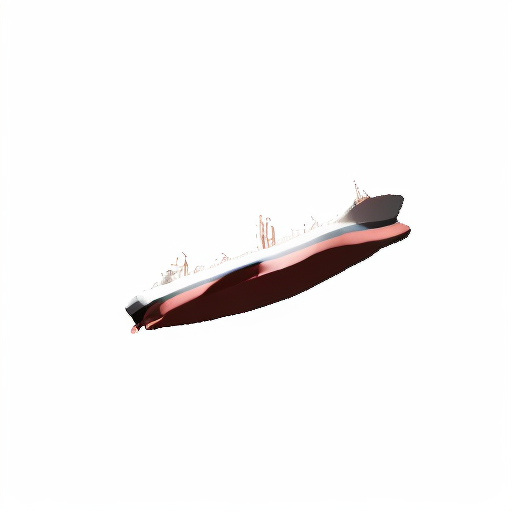} 
  &   \includegraphics[width=0.22\linewidth,trim=10 10 10 10, clip]{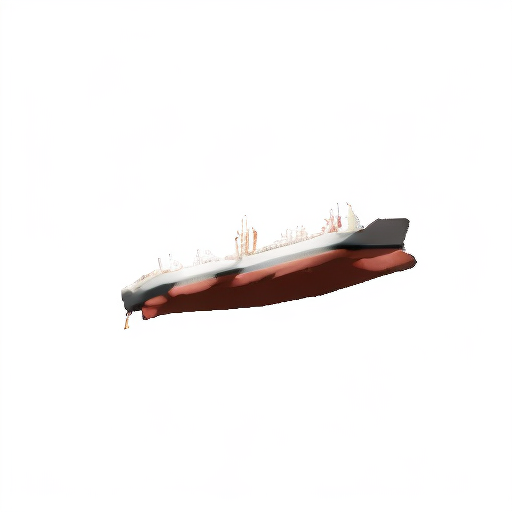}  \\  \vspace*{-15pt}

   \includegraphics[width=0.22\linewidth,trim=10 10 10 10, clip]{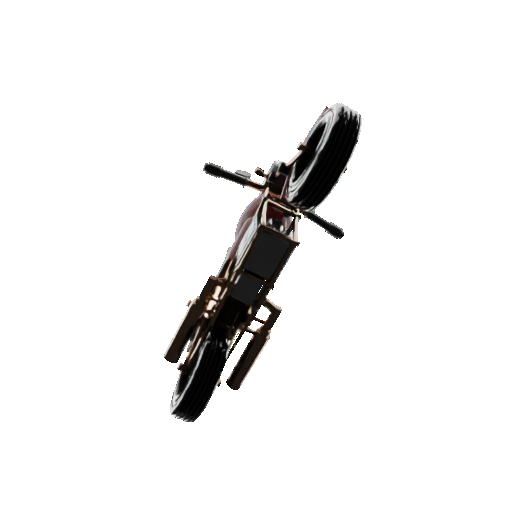} 
  & \includegraphics[width=0.22\linewidth,trim=10 10 10 10, clip]{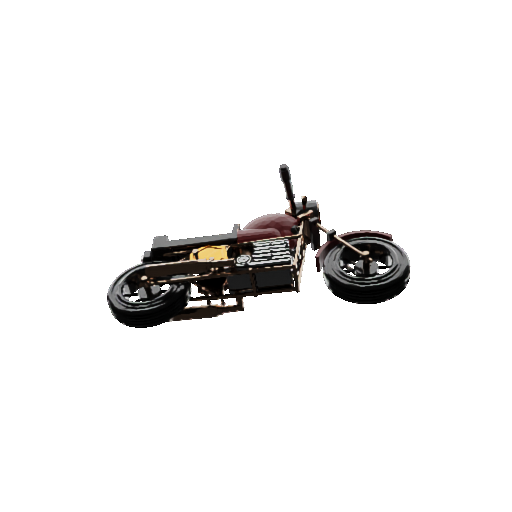} 
  &   \includegraphics[width=0.22\linewidth,trim=10 10 10 10, clip]{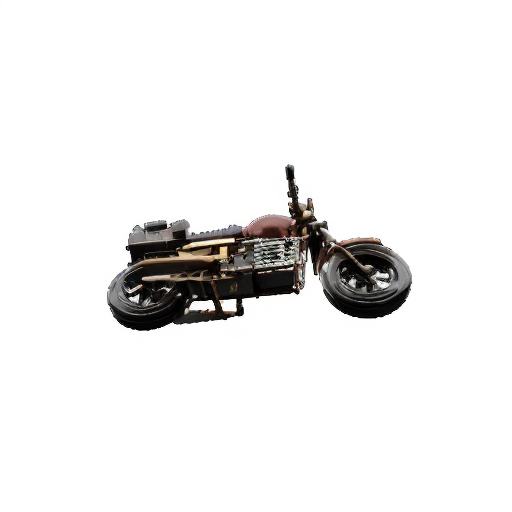} 
  &   \includegraphics[width=0.22\linewidth,trim=10 10 10 10, clip]{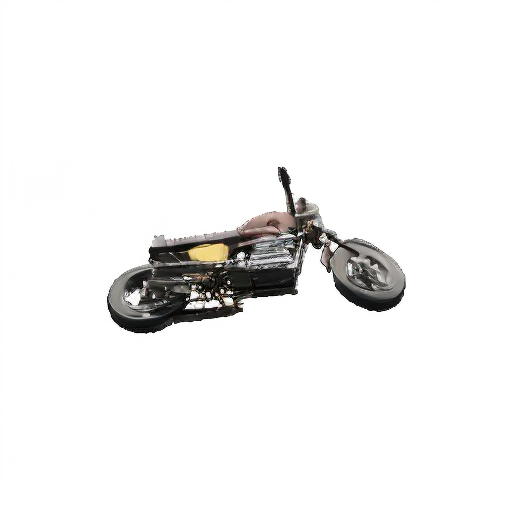}  \\   \vspace*{-15pt}

   \includegraphics[width=0.22\linewidth,trim=10 10 10 10, clip]{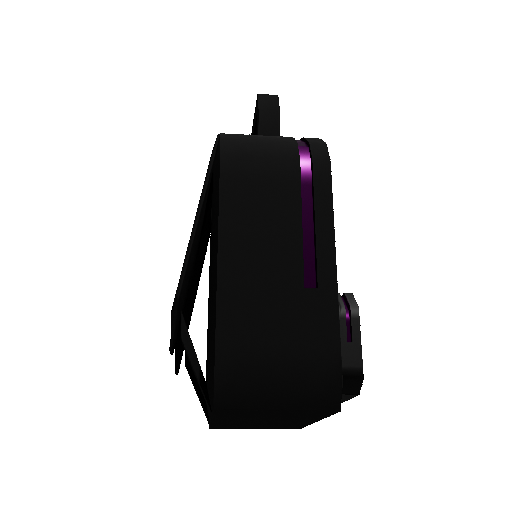} 
  & \includegraphics[width=0.22\linewidth,trim=10 10 10 10, clip]{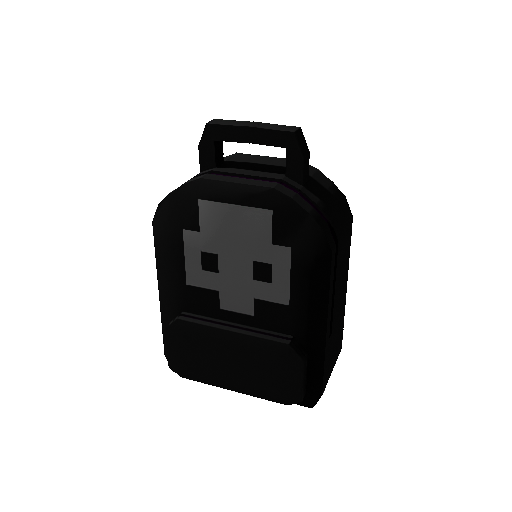} 
  &   \includegraphics[width=0.22\linewidth,trim=10 10 10 10, clip]{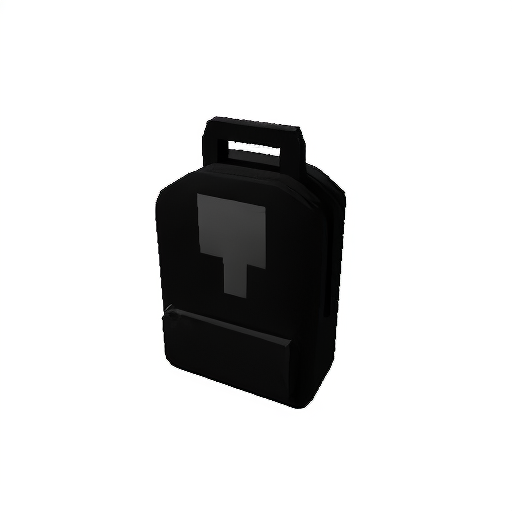} 
  &   \includegraphics[width=0.22\linewidth,trim=10 10 10 10, clip]{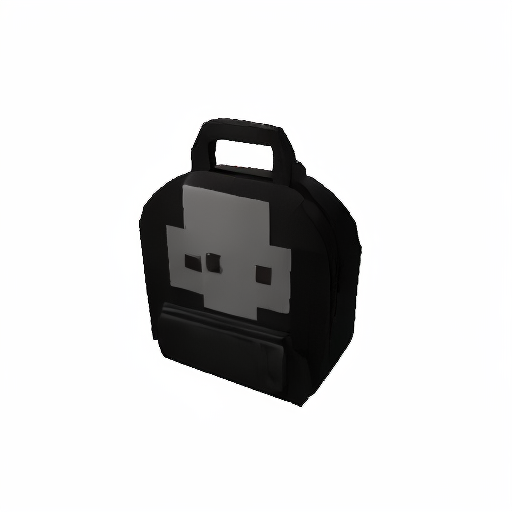} \\    \vspace*{-15pt}

     \includegraphics[width=0.22\linewidth,trim=10 10 10 10, clip]{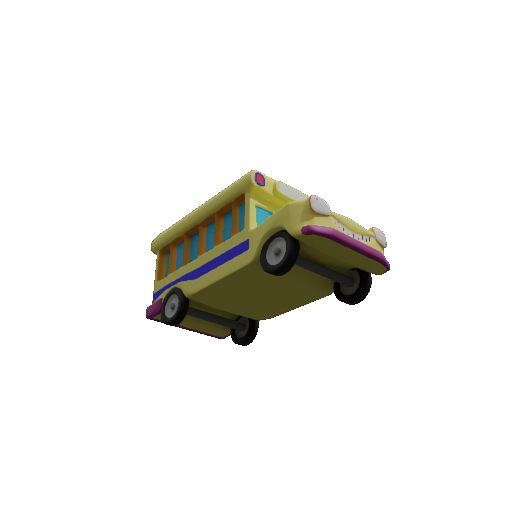} 
  & \includegraphics[width=0.22\linewidth,trim=10 10 10 10, clip]{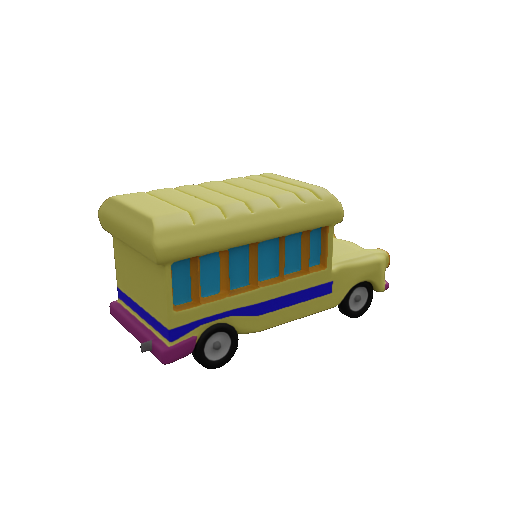} 
  &   \includegraphics[width=0.22\linewidth,trim=10 10 10 10, clip]{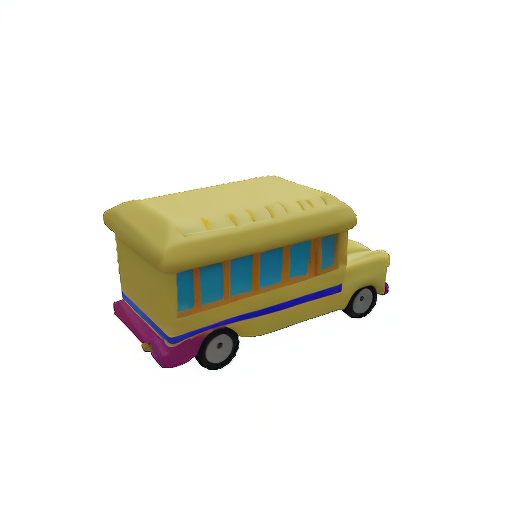} 
  &   \includegraphics[width=0.22\linewidth,trim=10 10 10 10, clip]{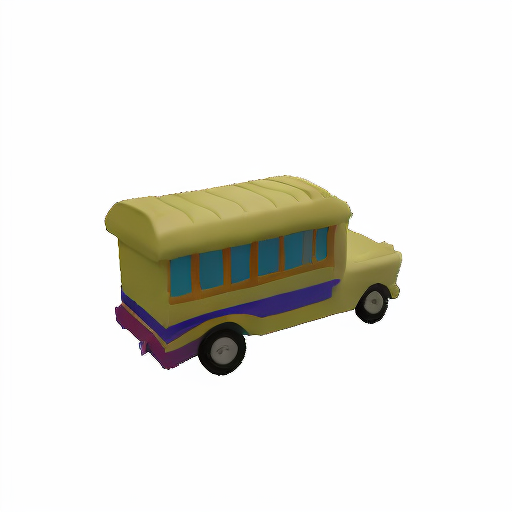} \\  \vspace*{-15pt}

  \includegraphics[width=0.22\linewidth]{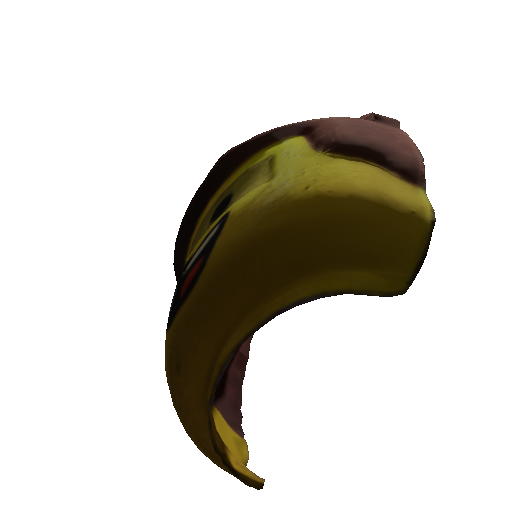} 
  & \includegraphics[width=0.22\linewidth]{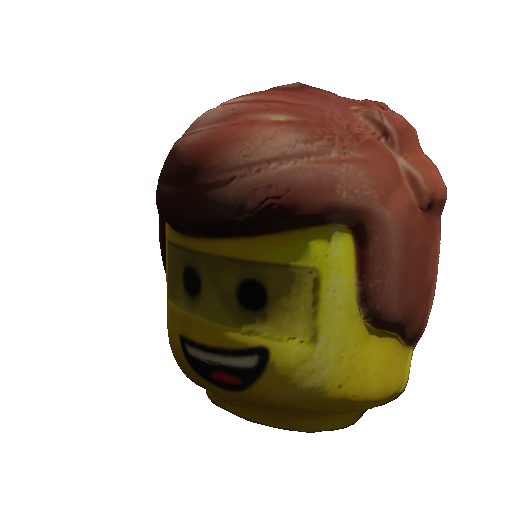} 
  &   \includegraphics[width=0.22\linewidth]{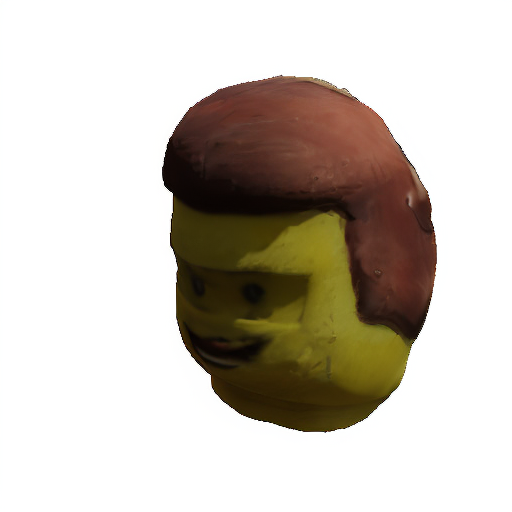} 
  &   \includegraphics[width=0.22\linewidth]{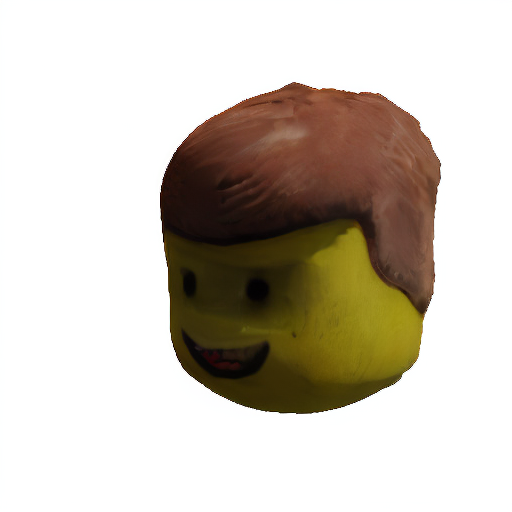} \\ [15pt]

Condition & Target & Linear & Geodesic  \\ [2pt]
\end{tabular}
\caption{
\textbf{Extended Qualitative Results} of PDG-FM (geodesic) and Linear-D2D-FM on Objaverse.  
}
\label{fig:more_pdgfm}
\end{figure}

\paragraph{Extended PDG-FM Qualitative Results}
Fig.~\ref{fig:more_pdgfm} presents additional comparisons between our Probability Density Geodesic FM (PDG-FM) and linear interpolation baselines. 
Visual analysis shows that our geodesic Data-to-Data FM (PDG-FM) preserves finer details across viewpoint changes, as geodesic interpolants maintain manifold consistency throughout transformations. 
These observations align with our quantitative findings in the main text, demonstrating the benefits of manifold-aware interpolation.

\end{document}